%% file: mmaml.tex
\documentclass{article}

\usepackage[final]{neurips_2019}

\setcitestyle{square,numbers,comma,sort&compress}

\usepackage[pagebackref=true,breaklinks=true,letterpaper=true,colorlinks,bookmarks=false,citecolor=blue]{hyperref}

\usepackage[utf8]{inputenc} 
\usepackage[T1]{fontenc}    
\usepackage{url}            
\usepackage{booktabs}       
\usepackage{amsfonts}       
\usepackage{nicefrac}       
\usepackage{microtype}      

\usepackage{graphicx}
\usepackage{subfigure}
\usepackage{amsmath}
\usepackage{amssymb}
\usepackage{caption}
\usepackage{courier}
\usepackage{dsfont}
\usepackage{mathtools}
\usepackage{dsfont}
\usepackage{multirow}
\usepackage{tabularx}
\usepackage{arydshln}
\usepackage{pifont}
\usepackage{algorithm}
\usepackage{algorithmic}
\usepackage{todonotes}
\usepackage{setspace}
\usepackage{wrapfig}
\usepackage{lipsum}

\usepackage{enumitem}
\setlist[itemize]{leftmargin=5mm, nolistsep}

\usepackage{color}
\usepackage{xcolor}

\input{macro.tex}

\title{\mytitle}

\author{
Risto Vuorio$^{\ast1}$ \hskip2em Shao-Hua Sun$^{\ast2}$ \hskip2em Hexiang Hu$^2$ \hskip2em Joseph J. Lim$^2$ \\
$^1$University of Michigan \hskip2em $^2$University of Southern California
\\ {\small \texttt{vuoristo@gmail.com}} \hskip2em {\small \texttt{\{shaohuas, hexiangh, limjj\}@usc.edu}}
}

\begin{document}

\maketitle

\begin{abstract}
\input{text/abstract.tex}
\end{abstract}


\input{text/introduction.tex}

\input{text/related_work.tex}

\input{text/problem.tex}

\input{text/method.tex}

\input{text/experiment.tex}

\input{text/conclusion.tex}

\input{text/acknowledgment.tex}

\clearpage
\small

\bibliography{mmaml}
\bibliographystyle{plain}

\clearpage
\input{text/appendix.tex}

\end{document}

%% file: macro.tex
\newcommand{\mytitle}{Multimodal Model-Agnostic Meta-Learning via Task-Aware Modulation}

\newcommand{\sun}[1]{{\color{blue}{\small\bf\sf [Sun: #1]}}}
\newcommand{\risto}[1]{{\color{red}{\small\bf\sf [Risto: #1]}}}
\newcommand{\hx}[1]{{\color{green}{\small\bf\sf [HX: #1]}}}
\newcommand{\joseph}[1]{{\color{cyan}{\small\bf\sf [Joseph: #1]}}}
\newcommand{\Skip}[1]{}

\newcommand{\ie}{\textit{i}.\textit{e}.\ }
\newcommand{\eg}{\textit{e}.\textit{g}.\ }

\newcommand{\myfig}[1]{Figure \ref{#1}}
\newcommand{\mytable}[1]{Table \ref{#1}}

\newcommand{\dotieconcat}[2]{
  \text{\raisebox{.8ex}{$\smallfrown$}}%
}

\def\aligntop#1{\vtop{\null\hbox{#1}}}

\newcommand\blfootnote[1]{%
  \begingroup
  \renewcommand\thefootnote{}\footnote{#1}%
  \addtocounter{footnote}{-1}%
  \endgroup
}

%% file: text/abstract.tex
Model-agnostic meta-learners aim to acquire meta-learned parameters from similar tasks to adapt to novel tasks from the same distribution with few gradient updates. With the flexibility in the choice of models, those frameworks demonstrate appealing performance on a variety of domains such as few-shot image classification and reinforcement learning. However, one important limitation of such frameworks is that they seek a common initialization shared across the entire task distribution, substantially limiting the diversity of the task distributions that they are able to learn from. In this paper, we augment MAML~\citep{finn_model-agnostic_2017} with the capability to identify the mode of tasks sampled from a multimodal task distribution and adapt quickly through gradient updates. Specifically, we propose a multimodal MAML (MMAML) framework, which is able to modulate its meta-learned prior parameters according to the identified mode, allowing more efficient fast adaptation. We evaluate the proposed model on a diverse set of few-shot learning tasks, including regression, image classification, and reinforcement learning.
The results not only demonstrate the effectiveness of our model in modulating the meta-learned prior in response to the characteristics of tasks but also show that training on a multimodal distribution can produce an improvement over unimodal training. 
The code for this project is publicly available at \url{https://vuoristo.github.io/MMAML}.
\blfootnote{$^\ast$Contributed equally.}

%% file: text/introduction.tex
\section{Introduction}
\label{sec:intro}
 
Humans make effective use of prior knowledge to acquire new skills rapidly.
When the skill of interest is related to a wide range of skills that one have mastered before, 
we can recall relevant knowledge of prior skills and exploit them to accelerate the new skill acquisition procedure.
For example, imagine that we are learning a novel snowboarding trick with knowledge of basic skills about snowboarding, skiing, and skateboarding.
We accomplish this feat quickly by exploiting our basic snowboarding knowledge together with inspiration from our skiing and skateboarding experience.

\Skip{
Humans make effective use of prior knowledge to acquire new skills rapidly.
Even when we have mastered a large set of skills before, 
we can often efficiently identify the relevant knowledge about the target skill (via preliminary explorations) and exploit them to accelerate the new skill acquisition procedure.
For example, imagine that we are learning a new sport, \ie, snowboarding, it is difficult if we do not have any knowledge about ``snowboarding'' is. Exploring snowboarding in person and watch the replay can significantly improve our understanding towards the characteristic of this target skill, so that we can learn faster about it via transferring relevant prior knowledge such as skateboarding.
}


Can machines likewise quickly master a novel skill based on a variety of related skills they have already acquired?
Recent advances in meta-learning~\cite{vinyals2016matching,finn_meta-learning_2017,duan2016rl} have attempted to tackle this problem. They offer machines a way to rapidly adapt to a new task using few samples by first learning an internal representation that matches similar tasks. 
Such representations can be learned by considering a distribution over similar tasks as the training data distribution.
Model-based (\ie RNN-based) meta-learning approaches~\citep{duan2016rl, wang2016learning, munkhdalai17a, mishra2018a} propose to recognize the task identity from a few sample data, use the task identity to adjust a model's state (\eg RNN's internal state or an external memory) and make the appropriate predictions with the adjusted model.
Those methods demonstrate good performance at the expense of having to hand-design architectures, yet the optimal strategy of designing a meta-learner for arbitrary tasks may not always be obvious to humans.
On the other hand, model-agnostic meta-learning frameworks~\citep{finn_model-agnostic_2017, finn_probabilistic_2018, kim_bayesian_2018, lee_gradient-based_2018, grant_recasting_2018, nichol2018reptile, rusu2018metalearning, rothfuss2019promp}
seek an initialization of model parameters that a small number of gradient updates will lead to superior performance on a new task.
With the flexibility in the model choices, these frameworks demonstrate appealing performance on a variety of domains, including regression, image classification, and reinforcement learning.

While most of the existing model-agnostic meta-learners rely on a single initialization,
different tasks sampled from a complex task distributions can require substantially different parameters, making it difficult to find a single initialization that is close to all target parameters.
If the task distribution is multimodal with disjoint and far apart modes (\eg snowboarding, skiing), one can imagine that a set of separate meta-learners with each covering one mode could better master the full distribution.
However, associating each task with one of the meta-learners not only requires additional task identity information, which is often not available or could be ambiguous when the modes are not clearly disjoint, 
but also disables transferring knowledge across different modes of the task distribution.
To overcome this issue, we aim to develop a meta-learner that is able to acquire mode-specific prior parameters and adapt quickly given tasks sampled from a multimodal task distribution.

To this end, we leverage the strengths of the two main lines of existing meta-learning techniques: model-based and model-agnostic meta-learning.
Specifically, we propose to augment MAML~\cite{finn_model-agnostic_2017} with the capability of generalizing across a multimodal task distribution.
Instead of learning a single initialization point in the parameter space, we propose to first compute the task identity of a sampled task by examining task related data samples.
Given the estimated task identity, our model then performs modulation to condition the meta-learned initialization on the inferred task mode.
Then, with these modulated parameters as the initialization, a few steps of gradient-based adaptation are performed towards the target task to progressively improve its performance.
An illustration of our proposed framework is shown in \myfig{fig:model}.

To investigate whether our method can acquire meta-learned prior parameters by learning tasks sampled from multimodal task distributions, we design and conduct experiments on a variety of domains, including regression, image classification, and reinforcement learning.
The results demonstrate the effectiveness of our approach against other systems.
A further analysis has also shown that our method learns to identify task modes without extra supervision. 

The main contributions of this paper are three-fold as follows:
\begin{itemize}

\item We identify and empirically demonstrate the limitation of having to rely on a single initialization in a family of widely used model-agnostic meta-learners.

\item We propose a framework together with an algorithm to address this limitation.
Specifically, it generates a set of meta-learned prior parameters and adapts quickly given tasks from a multimodal task distribution leveraging both model-based and model-agnostic meta-learning.

\item We design a set of multimodal meta-learning problems and demonstrate that our model offers a better generalization ability in a variety of domains, including regression, image classification, and reinforcement learning.

\end{itemize}

%% file: text/related_work.tex

\section{Related Work}
\label{sec:related_work}

The idea of empowering the machines with the capability of \textit{learning to learn}~\cite{thrun2012learning} has been widely explored by the machine learning community.
To improve the efficiency of handcrafted optimizers, a flurry of recent works has focused on learning to optimize a learner model.
Pioneered by~\cite{schmidhuber:1987:srl, bengio1992optimization}, optimization algorithms with learned parameters have been proposed, enabling the automatic exploitation of the structure of learning problems. 
From a reinforcement learning perspective, \cite{li_learning_2016} represents an optimization algorithm as a learning policy.
\cite{andrychowicz_learning_2016} trains LSTM optimizers to learn update rules from the gradient history, and \cite{ravi_optimization_2017} trains a meta-learner LSTM to update a learner's parameters. Similar approach for continual learning is explored in~\cite{vuorio2018meta}.

Recently, investigating how we can replicate the ability of humans to learn new concepts from one or a few instances, known as \textit{few-shot learning}, has drawn people's attention due to its broad applicability to different fields.
To classify images with few examples, 
metric-based meta-learning frameworks have been proposed~\cite{koch2015siamese, vinyals2016matching, snell2017prototypical, shyam2017attentive, Sung_2018_CVPR, oreshkin_tadam:_2018, chen2018a}, which strive to learn a metric or distance function that can be used to compare two different samples effectively. Recent works along this line~\cite{oreshkin_tadam:_2018,YeHZS2018Learning,lee2018gradient} share a conceptually similar idea with us and seek to perform task-specific adaptation with different type transformations. Due to the limited space, we defer the detailed discussion to the supplementary material.
While impressive results have been shown, it is nontrivial to adopt them for complex tasks such as acquiring robotic skills using reinforcement learning~\cite{hu2018synthesized, LillicrapHPHETS15, kalashnikov18a, Rajeswaran18, gu2017deep, haarnoja18b, lee2019composing}.

On the other hand, instead of learning a metric, model-based (\ie RNN-based) meta-learning models learn to adjust model states (\eg a state of an RNN~\cite{mishra2018a, duan2016rl, wang2018prefrontal} or external memory~\cite{santoro_meta-learning_2016, munkhdalai17a}) using a training dataset and output the parameters of a learned model or the predictions given test inputs.
While these methods have the capacity to learn any mapping from datasets and test samples to their labels, 
they could suffer from overfitting and show limited generalization ability~\cite{finn_meta-learning_2017}.

Model-agnostic meta-learners~\cite{finn_model-agnostic_2017, finn_probabilistic_2018, kim_bayesian_2018, lee_gradient-based_2018, grant_recasting_2018, nichol2018reptile, rusu2018metalearning, rothfuss2019promp} are agnostic to concrete model configurations. Specifically, they aim to learn a parameter initialization under a certain task distribution, that aims to provide a favorable inductive bias for fast gradient-based adaptation.
With its model agnostic nature, appealing results have been shown on a variety of learning problems.
However, assuming tasks are sampled from a concentrated distribution and pursuing a common initialization to all tasks can substantially limit the performance of such methods on multimodal task distributions where the center in the task space becomes ambiguous.

In this paper, we aim to develop a more powerful model-agnostic meta-learning framework which is able to deal with complex multimodal task distributions.
To this end, we propose a framework, which first identifies the mode of sampled tasks, similar to model-based meta-learning approaches, and then it modulates the meta-learned prior parameters to make the model better fit to the identified mode.
Finally, the model is fine-tuned on the target task rapidly through gradient steps.

%% file: text/problem.tex

\section{Preliminaries}
\label{sec:preliminaries}

The goal of meta-learning is to quickly learn task-specific functions that map between input data and the desired output $\left(x_k, y_k\right)^{\mathrm{K}_t}_{k=1}$ for different tasks $t$, where the number of data $\mathrm{K}_t$ is small. A task is defined by the underlying data generating distribution $\mathcal{P}(\mathrm{X})$ and a conditional probability $\mathcal{P}_t(\mathrm{Y} \mid \mathrm{X})$. 
For instance, we consider five-way image classification tasks with $x_k$ to be images and $y_k$ to be the corresponding labels, sampled from a task distribution.
The data generating distribution is unimodal if it contains classification tasks that belong to a single input and label domain (\eg {classifying different combination of digits}).
A multimodal counterpart therefore contains classification tasks from multiple different input and label domains (\eg {classifying digits vs. classifying birds}).
We denote the later distribution of tasks to be the \emph{multimodal task distribution}.

In this paper, we aim to rapidly adapt to a novel task sampled
from a multimodal task distribution.
We consider a target dataset $\mathcal{D}$ consisting of
tasks sampled from a multimodal distribution.
The dataset is split into meta-training and meta-testing sets, 
which are further divided into task-specific training $\mathcal{D}^{\text{train}}_{\mathcal{T}}$ and validation
$\mathcal{D}^{\text{val}}_{\mathcal{T}}$ sets. A meta-learner learns about
the underlying structure of the task distribution through training on the
meta-training set and is evaluated on meta-testing set. 

Our work builds upon Model-Agnostic Meta-Learning (MAML)
algorithm~\citep{finn_model-agnostic_2017}.
MAML seeks an initialization of parameters $\theta$ for
a meta-learner such that it can be optimized towards a new task with a small number of gradient
steps minimizing the task-specific objectives on the training data 
$\mathcal{D}^{\text{train}}_{\mathcal{T}}$,
with the adapted parameters generalize well to the validation
data $\mathcal{D}^{\text{val}}_{\mathcal{T}}$.
The initialization of the parameters is trained by sampling
mini-batches of tasks from $\mathcal{D}$, computing
the adapted parameters for all $\mathcal{D}^{\text{train}}_{\mathcal{T}}$ in the batch,
evaluating adapted parameters to compute the validation losses on the $\mathcal{D}^{\text{val}}_{\mathcal{T}}$ and finally update the initial parameters $\theta$ using the gradients from the validation losses.

%% file: text/method.tex
\section{Method}
\label{sec:method}

\input{text/model_and_algorithm.tex}

Our goal is to develop a framework to quickly master a novel task from \textit{a multimodal task distribution}. We call the proposed framework Multimodal Model-Agnostic Meta-Learning (MMAML).
The main idea of MMAML is to leverage two complementary neural networks to quickly adapt to a novel task. First, a network called the modulation network predicts the identity of the mode of a task. Then the predicted mode identity is used as an input by a second network called the task network, which is further adapted to the task using gradient-based optimization. Specifically, the modulation network accesses data points from the target task and produces a set of task-specific parameters to modulate the meta-learned prior parameters of the task network. 
Finally, the modulated task network (but not the task-specific parameters from modulation network) is further adapted to target task through gradient-based optimization.
A conceptual illustration can be found in \myfig{fig:model}.

In the rest of this section, we introduce our modulation network and a variety of modulation operators in section~\ref{subsec:modulation_network}. Then we describe our task network and the training details for MMAML in section~\ref{subsec:task_network}.

\subsection{Modulation Network}
\label{subsec:modulation_network}

As mentioned above, modulation network is responsible for identifying the mode of a sampled task, and generate a set of parameters specific to the task. To achieve this, it first takes the given $K$ data points and their labels $\{x_k, y_k\}_{k=1,...,K}$ as input to the task encoder $f$ and produces an embedding vector $\upsilon$
that encodes the characteristics of a task:
\begin{equation}
    \upsilon = h\Big(\{\left(x_k, y_k\right) \mid {k=1,\cdots,K} \}; \;\omega_h\Big)
\end{equation}
Then the task-specific parameters $\tau$ are computed based on the encoded task embedding vector $\upsilon$, 
which is further used to modulate the meta-learned prior parameters of the
task network. 
The task network (introduced later at Section~\ref{subsec:task_network}) can be an arbitrarily parameterized function, with multiple building blocks (or layers) such as deep convolutional networks~\cite{he2016deep}, or multi-layer recurrent networks~\cite{peters2018deep}.
To modulate the parameters of each block in the task network as good initialization for solving the target task,
we apply block-wise transformations to scale and shift the output activation of each hidden unit in the network
(\ie the output of a channel of a convolutional layer or a neuron of a fully-connected layer).
Specifically, the modulation network produces the modulation vectors for each block $i$, denoted as 
\begin{equation}
    \tau_i = g_i\left(\upsilon; \omega_{g}\right), \text{where \;} i = 1, \cdots, N,
\end{equation}
where $N$ is the number of blocks in the task network. We formalize the procedure of applying modulation as: $\phi_i = \theta_i \odot \tau_i$,
where $\phi_i$ is the modulated prior parameters for the task network, 
and $\odot$ represents a general modulation operator. 
We investigate some representative modulation operations
including attention-based (softmax) modulation~\citep{mnih2014recurrent,vaswani_attention_2017} 
and feature-wise linear modulation (FiLM)~\citep{perez_film:_2017, park2019semantic, huh2019feedback}. 
We empirically observe that FiLM performs better and more stable than attention-based modulation (see Section~\ref{sec:experiments} for details), 
and therefore use FiLM as default operator for modulation. The details of these modulation operators can be found in the supplementary material.

\subsection{Task Network} 
\label{subsec:task_network}

The parameters of each block of the task network are modulated using the task-specific parameters $\tau = \{ \tau_i \mid i = 1, \cdots, N \}$ generated by the modulation network, which can generate a mode-aware initialization in the parameter space $f(x ; \theta, \tau)$. 
After the modulation step, few steps of gradient descent are performed on the meta-learned prior parameters of the task network to further optimize the objective function for a target task $\mathcal{T}_{i}$. Note that the task-specific parameters $\tau_i$ are kept fixed and only the meta-learned prior parameters of the task network are updated. We describe the concrete procedure in the form of the pseudo-code as shown in Algorithm~\ref{alg:train}. The same procedure of modulation and gradient-based optimization is used both during meta-training and meta-testing time.



Detailed network architectures and training hyper-parameters are different by the domain of applications, we defer the complete details to the supplementary material.

%% file: text/model_and_algorithm.tex
\begin{figure*}[ttt!]

\begin{tabular}{cc}
\begin{minipage}{.45\textwidth}
    
    \centering
    \includegraphics[width=\textwidth]{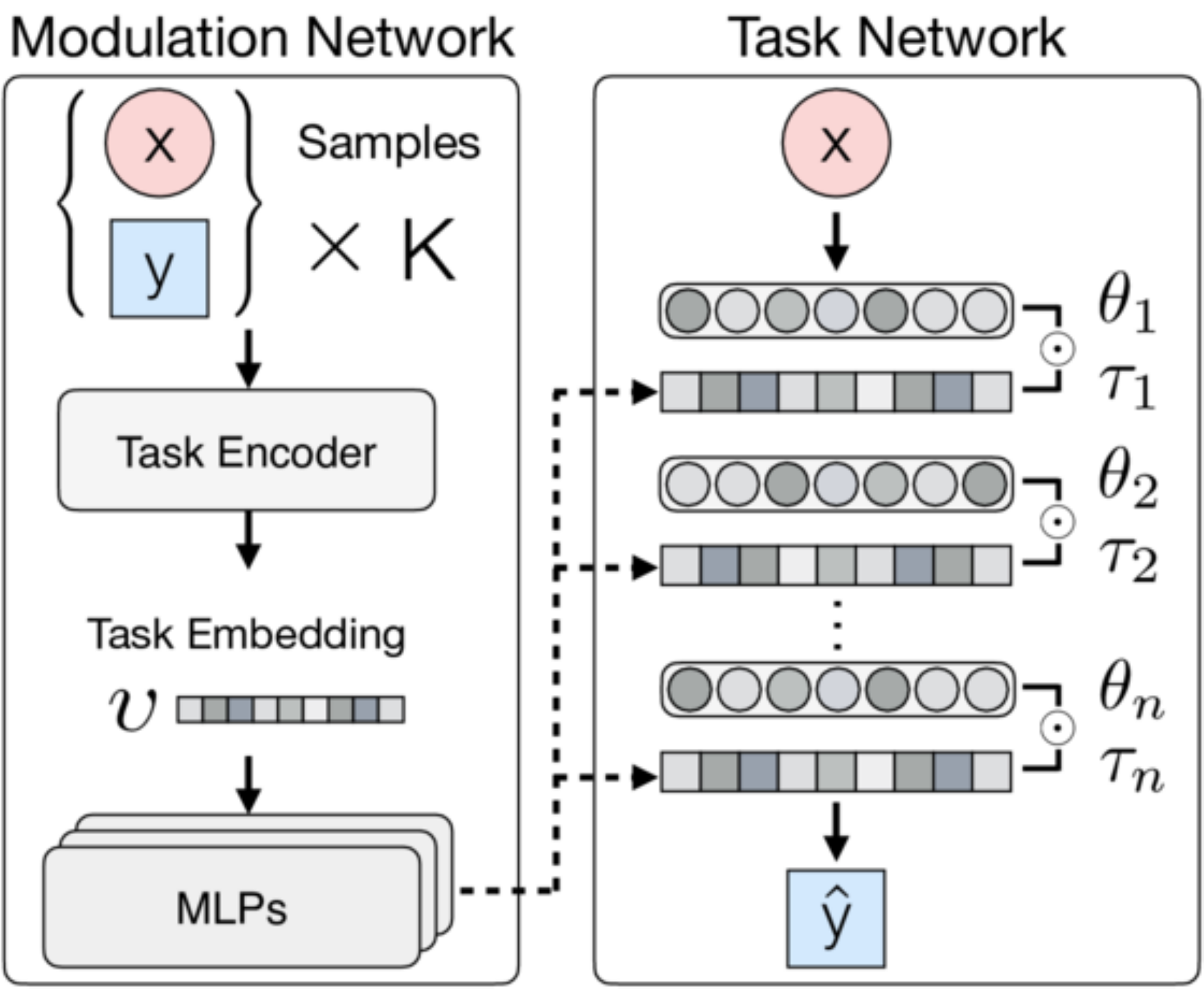}
    \captionof{figure}{
        \small \textbf{Model overview.} 
        The modulation network produces a task embedding $\upsilon$ , which is used to generate parameters $\{ \tau_i \}$ that modulates the task network.
        The task network adapts modulated parameters to fit to the target task.
    } \label{fig:model}

\end{minipage} &

\hfill

\scalebox{0.75}{
    \begin{minipage}{.64\textwidth}
    \begin{algorithm}[H]
    \captionof{algorithm}{\textsc{MMAML Meta-Training Procedure.}} \label{alg:train}
        \begin{spacing}{1.35}
        \begin{algorithmic}[1]
        \STATE {\bfseries Input:} Task distribution $P(\mathcal{T})$, Hyper-parameters $\alpha$ and $\beta$
    		\STATE Randomly initialize $\theta$ and $\omega$.
    		\WHILE{not DONE}
        		\STATE Sample batches of tasks $\mathcal{T}_j \sim P(\mathcal{T})$
        		\FOR{$\mathbf{all}$ j}
        		    \STATE Infer $\upsilon = h(\{x, y \}_K ; \omega_h)$ with K samples from  $\mathcal{D}^{\text{train}}_{\mathcal{T}_j}$.
        		    \STATE Generate parameters $\mathbf{\tau} = \{ g_i(\upsilon; \omega_g ) \mid i = 1, \cdots, N \}$ to modulate each block of the task network $f$.
        		    \STATE Evaluate $\nabla_{\theta} \mathcal{L}_{\mathcal{T}_j}(f(x ; \theta, \tau);\mathcal{D}^{\text{train}}_{\mathcal{T}_j})$ 
        		    w.r.t the K samples
        		    \STATE Compute adapted parameter with gradient descent: \\ $\theta_{\mathcal{T}_j}' = \theta - \alpha \nabla_{\theta} \mathcal{L}_{\mathcal{T}_j}\big(f(x ; \theta, \tau); \mathcal{D}^{\text{train}}_{\mathcal{T}_j}\big)$
        		\ENDFOR
        		\STATE Update $\theta$ with $\beta \nabla_{\theta} \sum_{T_j \sim P(\mathcal{T})} \mathcal{L}_{\mathcal{T}_j}\big(f(x ; \theta', \tau); \mathcal{D}^{\text{val}}_{\mathcal{T}_j}\big)$
        		\STATE Update $\omega_g$ with $\beta \nabla_{\omega_g} \sum_{T_j \sim P(\mathcal{T})} \mathcal{L}_{\mathcal{T}_j}\big(f(x ; \theta', \tau); \mathcal{D}^{\text{val}}_{\mathcal{T}_j}\big)$
        		\STATE Update $\omega_h$ with $\beta \nabla_{\omega_h} \sum_{T_j \sim P(\mathcal{T})} \mathcal{L}_{\mathcal{T}_j}\big(f(x ; \theta', \tau); \mathcal{D}^{\text{val}}_{\mathcal{T}_j}\big)$
    		\ENDWHILE  
    \end{algorithmic}
    \end{spacing}
    
    \end{algorithm}
    
    \end{minipage}
}

\end{tabular}

\end{figure*}

%% file: text/experiment.tex
\section{Experiments}
\label{sec:experiments}

We evaluate our method  (MMAML) and baselines in a variety of domains including regression, image classification, and reinforcement learning, under the multimodal task distributions. 
\Skip{
\subsection{Baselines}
\label{sec:baseline}
}
We consider the following model-agnostic meta-learning baselines:

\begin{itemize}[noitemsep]
\item \textbf{MAML~\cite{finn_model-agnostic_2017}} represents the family of model-agnostic meta-learners. The architecture of MAML on each individual domain is designed to be the same as task network in MMAML.

\item \textbf{Multi-MAML} consists of $M$ (the number of modes) MAML models and each of them is specifically trained on the tasks sampled from a single mode.
The performance of this baseline is evaluated by choosing models based on \textit{ground-truth task-mode labels}.
This baseline can be viewed as the upper-bound of performance for MAML.
If it outperforms MAML, it indicates that MAML's performance is degenerated due to the multimodality of task distributions.
Note that directly comparing the other algorithms to Multi-MAML is not fair as it uses additional information which is not available in real world scenarios.

\end{itemize}


Note that we aim to develop a general model-agnostic meta-learning framework and therefore the comparison to methods that achieved great performance on only an individual domain are omitted.
A more detailed discussion can be found in the supplementary material.

\input{text/experiment_regression.tex}

\input{text/experiment_few_shot.tex}

\input{text/experiment_rl.tex}

%% file: text/experiment_regression.tex
\subsection{Regression Experiments}
\label{sec:regression}

\begin{figure*}
	\centering
	\tabcolsep 2pt
	\small
	\begin{tabular}{ccccc}
		\multicolumn{5}{c}{\includegraphics[width=0.75\textwidth]{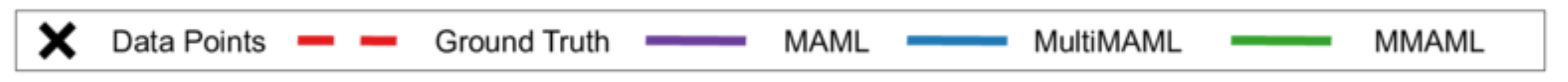}} \\
	    \scriptsize{\textbf{Sinusoidal}} & \scriptsize{\textbf{Linear}} &
	    \scriptsize{\textbf{Quadratic}}  & \scriptsize{\textbf{Transformed $\ell_1$ Norm}} & 
	    \scriptsize{\textbf{Tanh}} \\

		\includegraphics[width=0.185\textwidth,trim={0.6cm 0 0.2cm 0},clip]{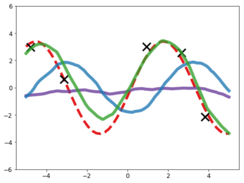} &
		\includegraphics[width=0.185\textwidth,trim={0.6cm 0 0.2cm 0},clip]{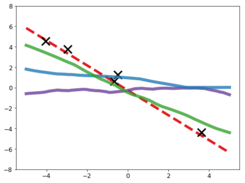} &
		\includegraphics[width=0.185\textwidth,trim={0.6cm 0 0.2cm 0},clip]{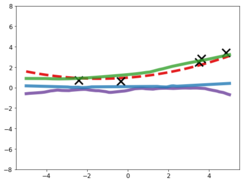} &
		\includegraphics[width=0.185\textwidth,trim={0.6cm 0 0.2cm 0},clip]{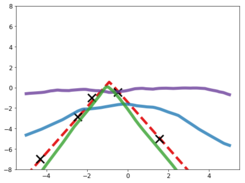} &
		\includegraphics[width=0.185\textwidth,trim={0.6cm 0 0.2cm 0},clip]{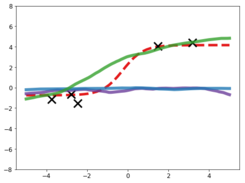}
		\\
		\multicolumn{5}{c}{\footnotesize \textbf{(a)} MMAML \textbf{post modulation} vs. other prior models} \\
		\includegraphics[width=0.185\textwidth,trim={0.6cm 0 0.2cm 0},clip]{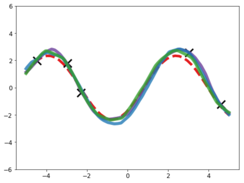} &
		\includegraphics[width=0.185\textwidth,trim={0.6cm 0 0.2cm 0},clip]{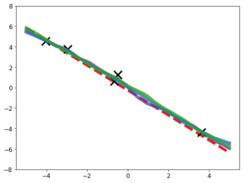} &
		\includegraphics[width=0.185\textwidth,trim={0.6cm 0 0.2cm 0},clip]{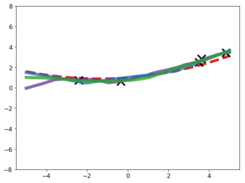} &		\includegraphics[width=0.185\textwidth,trim={0.6cm 0 0.2cm 0},clip]{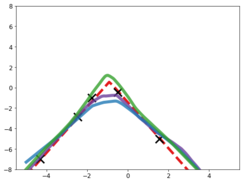} &
		\includegraphics[width=0.185\textwidth,trim={0.6cm 0 0.2cm 0},clip]{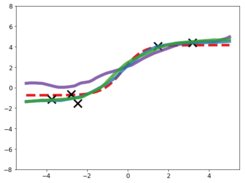} \\
		\multicolumn{5}{c}{\footnotesize \textbf{(b)} MMAML \textbf{post adaptation} vs. other posterior models} \\

	\end{tabular}
	
	\caption{
    	\small
	    Qualitative Visualization of Regression on Five-modes Simple Functions Dataset. 
	    \textbf{(a)}: We compare the predicted function shapes of modulated MMAML against the prior models of MAML and Multi-MAML, before gradient updates. Our model can fit the target function with limited observations and no gradient updates. 
	    \textbf{(b)}: The predicted function shapes after five steps of gradient updates, MMAML is qualitatively better. More visualizations in Supplementary Material.
	    \label{fig:regression}
	}
\end{figure*}

\input{text/table/regression_table.tex}

\noindent{\bf Setups.} We experiment with our models in multimodal few-shot regression. In our setup, five pairs of input/output data $\{x_k, y_k\}_{k=1,...,K}$ are sampled from a one dimensional function and provided to a learning model. The model is asked to predict $L$ output values $y_1^q, ..., y_L^q$ for input queries $x_1^q, ..., x_L^q$.
To construct the multimodal task distribution, we set up five different functions: sinusoidal, linear, quadratic, transformed $\ell_1$ norm, and hyperbolic tangent functions, and treat them as discrete task modes. We then evaluate three different task combinations with two functions, three functions and five functions in them. For each task, five pairs of data are sampled and Gaussian noise is added to the output value $y$, which further increases the difficulty of identifying which function generated the data. Please refer to the supplementary materials for details and parameters for regression experiments.


\noindent{\bf Baselines and Our Approach.} As mentioned before, we have MAML and Multi-MAML as two baseline methods, both with MLP task networks. Our method (MMAML) augments the task network with a modulation network. We choose to use an LSTM to serve as the modulation network due to its nature as good at handling sequential inputs and generate predictive outputs. Data points (sorted by $x$ value) are first input to this network to generate task-specific parameters that modulate the task network. The modulated task network is then further adapted using gradient-based optimization. Two variants of modulation operators -- softmax and FiLM are explored to be used in our approach. Additionally, to study the effectiveness of the LSTM model, we evaluate another baseline (referred to as the LSTM Learner) that uses the LSTM as the modulation network (with FiLM) but does not perform gradient-based updates. Please refer to the supplementary materials for concrete specification of each model.


\noindent{\bf Results.} The quantitative results are shown in \mytable{table:regression}. We observe that MAML has the highest error in all settings and that incorporating task identity (Multi-MAML) can improve over MAML significantly. This suggests that MAML degenerates under multimodal task distributions. The LSTM learner outperforms both MAML and Multi-MAML, showing that the sequence model can effectively tackle this regression task. MMAML improves over the LSTM learner significantly, which indicates that with a better initialization (produced by the modulation network), gradient-based optimization can lead to superior performance. Finally, since FiLM outperforms Softmax consistently in the regression experiments, we use it for as the modulation method in the rest of experiments.

We visualize the predicted function shapes of MAML, Multi-MAML and MMAML (with FiLM) in \myfig{fig:regression}. We observe that modulation can significantly modify the prediction of the initial network  to be close to the target function (see \myfig{fig:regression} (a)). The prediction is then further improved by gradient-based optimization (see \myfig{fig:regression} (b)). tSNE~\cite{maaten2008visualizing} visualization of the task embedding (\myfig{fig:tSNE}) shows that our embedding learns to separate the input data of different tasks, which can be seen as a evidence of the mode identification capability of MMAML.

%% file: text/table/regression_table.tex

\begin{table*}[t]
    \centering
    \small
    \caption{\small 
    Mean square error (MSE) on the \textbf{multimodal 5-shot regression} with 2, 3, and 5 modes. A Gaussian noise with $\mu=0$ and $\sigma=0.3$ is applied. Multi-MAML uses ground-truth task modes to select the corresponding MAML model. Our method (with FiLM modulation) outperforms other methods by a margin.}
    \scalebox{0.85}{
    \begin{tabular}{lcccccc} 
    \toprule
    \multirow{2}{*}{\textbf{Method}} & 
    \multicolumn{2}{c}{\textbf{2 Modes}} &
    \multicolumn{2}{c}{\textbf{3 Modes}} &
    \multicolumn{2}{c}{\textbf{5 Modes}} \\
    \cmidrule(lr){2-3}
    \cmidrule(lr){4-5}
    \cmidrule(lr){6-7}
    
    & \scriptsize{Post Modulation} & \scriptsize{Post Adaptation}
    & \scriptsize{Post Modulation} & \scriptsize{Post Adaptation}
    & \scriptsize{Post Modulation} & \scriptsize{Post Adaptation} \\
    \midrule
    MAML~\citep{finn_model-agnostic_2017} & - & 1.085 & - & 1.231 & - & 1.668 \\
    Multi-MAML   & - & 0.433 & - & 0.713 & - & 1.082 \\
    LSTM Learner &  0.362 & - & 0.548 & - & 0.898 & - \\
    \midrule
    \textbf{Ours:} MMAML (Softmax) & 1.548 & 0.361 & 2.213 & \textbf{0.444} & 2.421 & 0.939 \\
    \textbf{Ours:} MMAML (FiLM) & 2.421 & \textbf{0.336} & 1.923 & \textbf{0.444} & 2.166 & \textbf{0.868} \\
    \bottomrule
    \end{tabular}}
    \label{table:regression}
\end{table*}

%% file: text/experiment_few_shot.tex
\subsection{Image Classification}
\label{sec:fewshot}

\begin{figure}[t]
    \centering
    \begin{tabular}{cccc}
    \includegraphics[width=.22\textwidth]{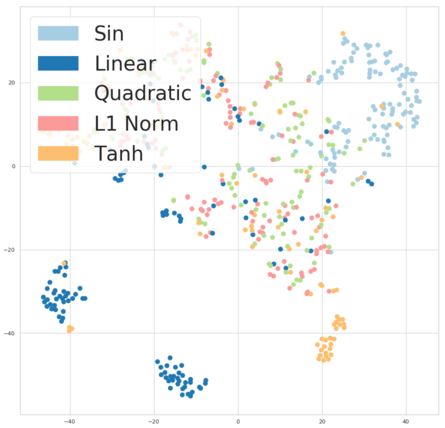} &
    \includegraphics[width=.22\textwidth]{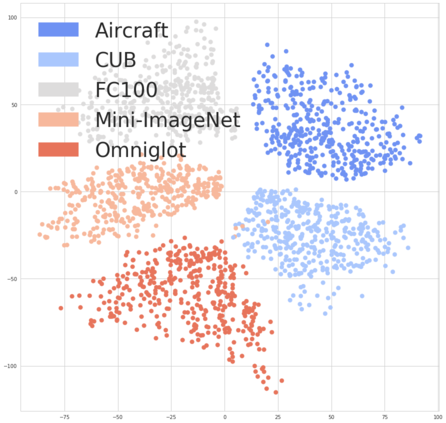} &
    \includegraphics[width=.22\textwidth]{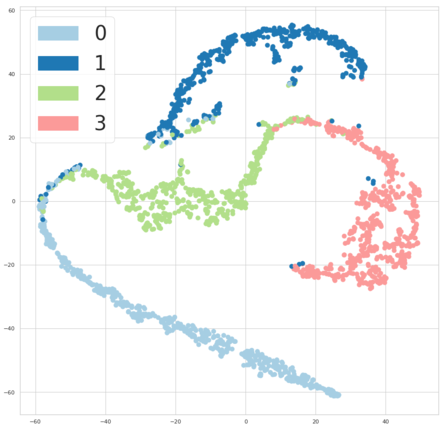} &
    \includegraphics[width=.22\textwidth]{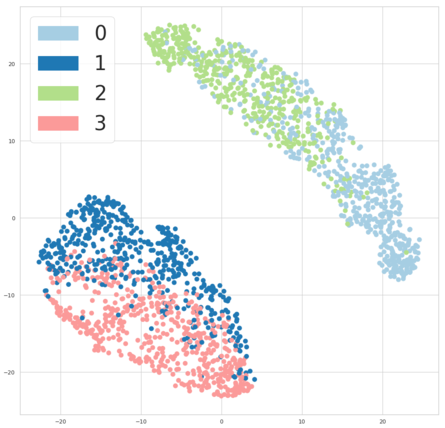} \\
    \small{(a) Regression} & (b) \small{Image classification} & 
    (c) \small{RL Reacher} & (d) \small{RL Point Mass} \\
    \end{tabular}
    
    \caption{ \small
        tSNE plots of the task embeddings produced by our model from randomly sampled tasks; marker color indicates different modes of a task distribution.
        The plots (b) and (d) reveal a clear clustering according to different task modes,
        which demonstrates that MMAML is able to identify the task from a few samples and produce a meaningful embedding $\upsilon$. 
        (a) Regression: the distance between modes aligns with the intuition of the similarity of functions (\eg a quadratic function can sometimes be similar to a sinusoidal or a linear function while a sinusoidal function is usually different from a linear function) 
        (b) Few-shot image classification: each dataset (\ie mode) forms its own cluster.
        (c-d) Reinforcement learning: The numbered clusters represent different modes of the task distribution. The tasks from different modes are clearly clustered together in the embedding space.
    }
    \label{fig:tSNE}
\end{figure}

\input{text/table/classification_table.tex}

\noindent{\bf Setup.} The task of few-shot image classification considers the problem of classifying images into $N$ classes with a small number ($K$) of labeled samples available (\ie $N$-way $K$-shot). 
To create a multimodal few-shot image classification task,
we combine multiple widely used datasets (\textsc{Omniglot}~\cite{omniglot}, \textsc{Mini-ImageNet}~\cite{ravi_optimization_2017},  \textsc{FC100}~\cite{oreshkin_tadam:_2018},
\textsc{CUB}~\cite{dataset-bird}, and \textsc{Aircraft}~\cite{dataset-aircraft})
to form a meta-dataset following the train/test splits used in the prior work, similar to~\cite{metadataset}.
The details of all the datasets can be found in the supplementary material.

We train models on the meta-datasets with two modes (\textsc{Omniglot} and \textsc{Mini-ImageNet}), three modes (\textsc{Omniglot}, \textsc{Mini-ImageNet}, and \textsc{FC100}), and five modes (all the five datasets).
We use a 4-layer convolutional network for both MAML and our task network.

\noindent{\bf Results.} The results are shown in \mytable{table:classification}, demonstrating that our method achieves better results against MAML and performs comparably to Multi-MAML.
The performance gap between ours and MAML becomes larger when the number of modes increases, suggesting our method can handle multimodal task distributions better than MAML.
Also, compared to Multi-MAML, our method achieves slightly better results partially because our method learns from all the datasets yet each Multi-MAML is likely to overfit to a single dataset with a smaller number of classes (\eg \textsc{Mini-ImageNet} and \textsc{FC100}).
This finding aligns with the current trend of meta-learning from multiple datasets~\cite{metadataset}.
The detailed performance on each dataset can be found in the supplementary material.

To gain insights to the task embeddings $\upsilon$ produced by our model,
we randomly sample 2000 5-mode 5-way 1-shot tasks and employ tSNE to visualize $\upsilon$ in \myfig{fig:tSNE} (b), showing that our task embedding network captures the relationship among modes, where each dataset forms an individual cluster.
This structure shows that our task encoder learns a reasonable task embedding space,
which allows the modulation network to modulate the parameters of the task network accordingly.

%% file: text/table/classification_table.tex
\begin{table*}[t]
    \centering
    \small
    \caption{\small Classification testing accuracies on the \textbf{multimodal few-shot image classification} with 2, 3, and 5 modes.
    Multi-MAML uses ground-truth dataset labels to select corresponding MAML models. Our method outperforms MAML and achieve comparable results with Multi-MAML in all the scenarios.}
    \scalebox{0.9}{
    \begin{tabular}{cccccccccc} 
    \toprule
    \textbf{Method \& Setup} & 
    \multicolumn{3}{c}{\textbf{2 Modes}} &
    \multicolumn{3}{c}{\textbf{3 Modes}} &
    \multicolumn{3}{c}{\textbf{5 Modes}}\\ 
    \cmidrule(lr){2-4}
    \cmidrule(lr){5-7}
    \cmidrule(lr){8-10}
    Way & \multicolumn{2}{c}{5-way} & 20-way & \multicolumn{2}{c}{5-way} & 20-way  & \multicolumn{2}{c}{5-way} & 20-way \\
    \cmidrule(lr){2-3}
    \cmidrule(lr){4-4}
    \cmidrule(lr){5-6}
    \cmidrule(lr){7-7}
    \cmidrule(lr){8-9}
    \cmidrule(lr){10-10}
    Shot & 1-shot & 5-shot & 1-shot & 1-shot & 5-shot & 1-shot & 1-shot & 5-shot & 1-shot \\
    
    \midrule
    
    MAML~\citep{finn_model-agnostic_2017} & 
    66.80\% & 77.79\% & 44.69\% & 54.55\% & 67.97\% & 28.22\% & 
    44.09\% & 54.41\% & 28.85\% \\

    Multi-MAML & 
    66.85\% & 73.07\% & \textbf{53.15}\% & 55.90\% & 62.20\% & \textbf{39.77}\% & 
    45.46\% & 55.92\% & 33.78\% \\

    \midrule
    MMAML (ours) & 
    \textbf{69.93}\% & \textbf{78.73}\% & 47.80\% & \textbf{57.47}\% & \textbf{70.15}\% & 36.27\% & 
    \textbf{49.06}\% & \textbf{60.83}\% & \textbf{33.97}\% \\

    \bottomrule
    \end{tabular}
	}
    \label{table:classification}
\end{table*}

%% file: text/experiment_rl.tex
\subsection{Reinforcement Learning}
\label{sec:RL}

\begin{figure*}[!htpb]
    \centering
    \begin{tabular}{cccc}
    
    \includegraphics[width=0.2\textwidth]{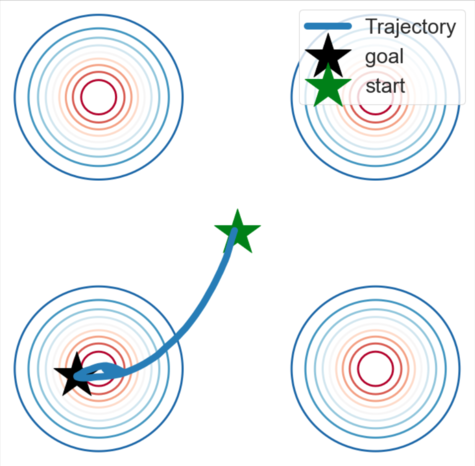} &
    \includegraphics[width=0.2\textwidth]{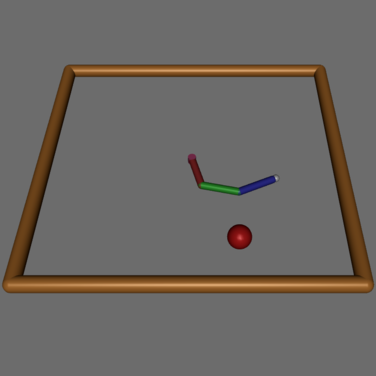} &
    \includegraphics[width=0.2\textwidth]{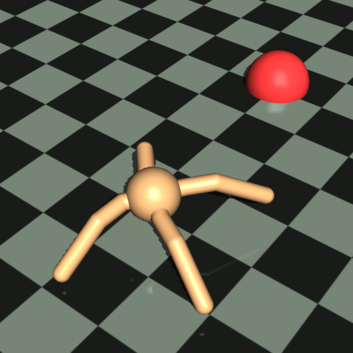} &
    \includegraphics[width=0.2\textwidth]{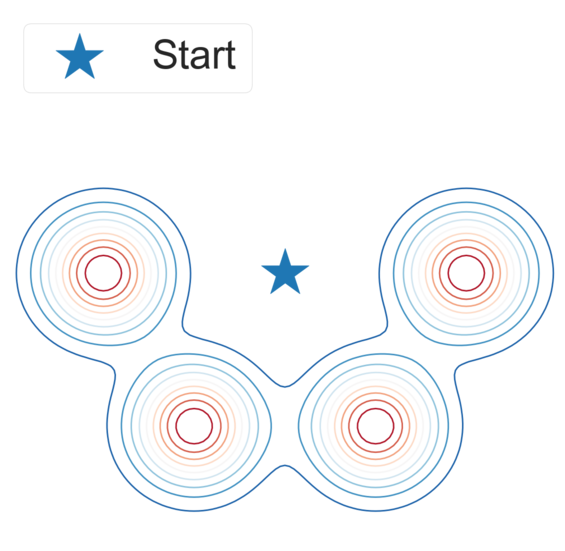}
    \\
    \small{(a) Point Mass} & 
    \small{(b) Reacher} & 
    \small{(c) Ant} & 
    \small{(d) Ant Goal Distribution}
    \end{tabular}
    
    \caption{ \small
        \textbf{RL environments}. Three environments are used to explore the capability of MMAML to adapt in multimodal task distributions in RL. In all of the environments the agent is tasked to reach a goal marked by a star of a sphere in the figures. The goals are sampled from a multimodal distribution in two or three dimensions depending on the environment. In \textsc{Point Mass} (a) the agent navigates a simple point mass agent in 2-dimensions. In \textsc{Reacher} (b) the agent controls a 3-link robot arm in 2-dimensions. In \textsc{Ant} (c) the agent controls four-legged ant robot and has to navigate to the goal. The goals are sampled from a 2-dimensional distribution presented in figure (d), while the agent itself is 3-dimensional.
        \label{fig:rl_example}
    }
\end{figure*}

\begin{figure}
    \centering
    \includegraphics[width=\textwidth]{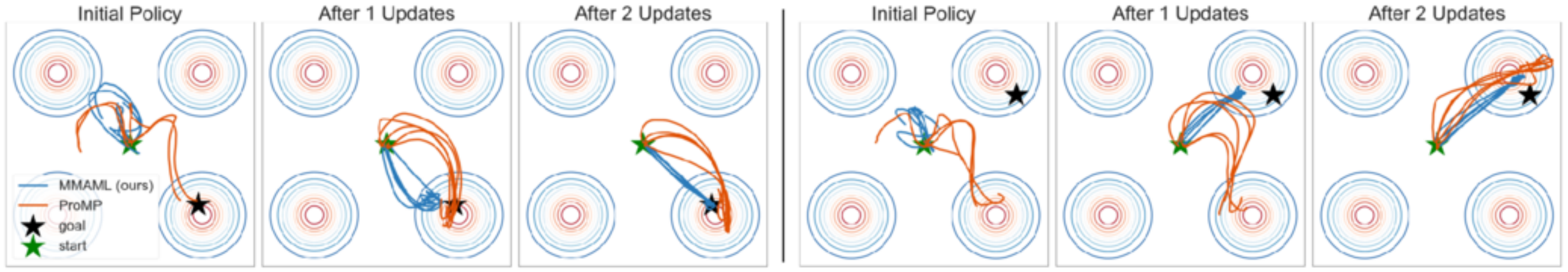}
    \caption{Visualizations of MMAML and ProMP trajectories in the 4-mode Point Mass 2D environment. Each trajectory originates in the green star. The contours present the multimodal goal distribution. Multiple trajectories are shown per each update step. 
    For each column:
    \textbf{the leftmost figure} depicts the initial exploratory trajectories without modulation or gradient adaptation applied. 
    \textbf{The middle figure} presents ProMP after one gradient adaptation step and MMAML after a gradient adaptation step and the modulation step, which are computed based on the same initial trajectories. 
    \textbf{The figure on the right} presents the methods after two gradient adaptation steps in addition to the MMAML modulation step.}
    \label{fig:rl_trajectories}
\end{figure}{}

\begin{figure}
    \centering
    \includegraphics[width=\textwidth]{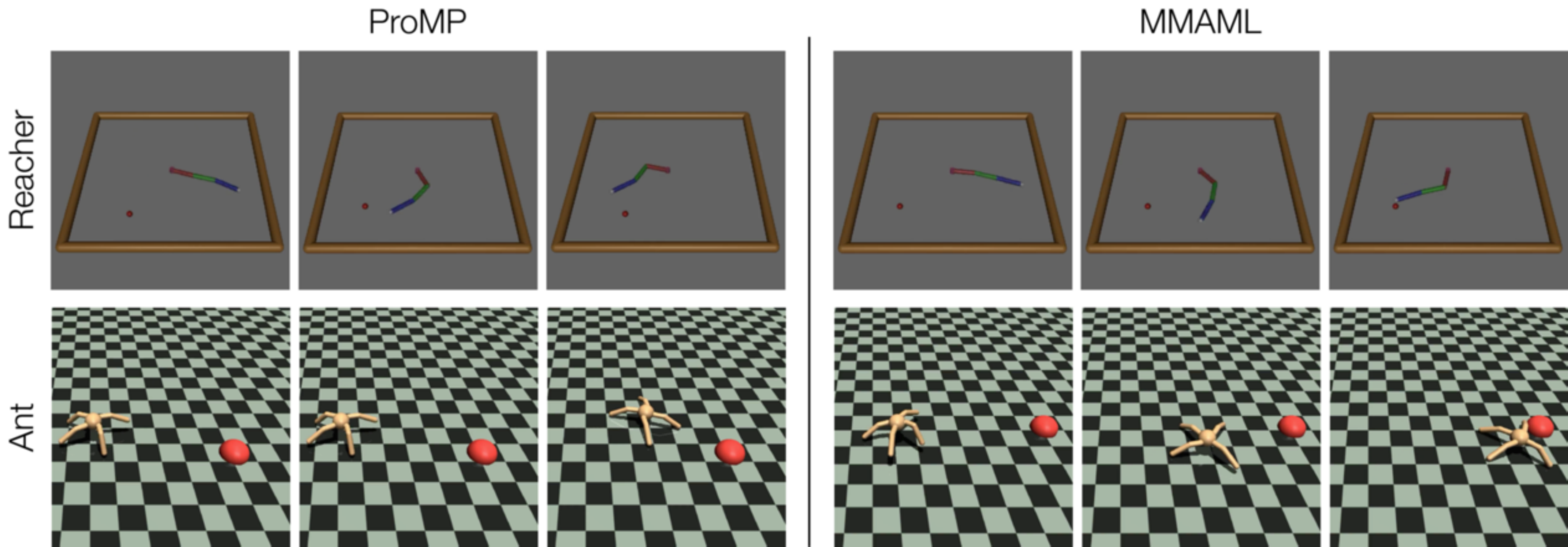}
    \caption{Visualizations of MMAML and ProMP trajectories in the \textsc{Ant} and \textsc{Reacher} environments. The figures represent randomly sampled trajectories after the modulation step and two gradient steps for \textsc{Reacher} and three for \textsc{Ant}. Each frame sequence represents a complete trajectory, with the beginning, middle and end of the trajectories captured by the left, middle and right frames respectively. Videos of the trained agents can be found at \url{https://vuoristo.github.io/MMAML/}.
    }
    \label{fig:rl_screenshots}
\end{figure}{}

\noindent{\bf Setup.}
Along with few-shot classification and regression, reinforcement learning (RL) has been a central problem where meta-learning has been studied \cite{schmidhuber1997shifting,schmidhuber1998reinforcement,wang2016learning,finn_model-agnostic_2017,mishra2018a,rothfuss2019promp}. 
Similarly to the other domains, the objective in meta-reinforcement learning (meta-RL) is to adapt to a novel task based on limited experience with the task. For RL problems, the inner loop updates of gradient-based meta-learning take the form of policy gradient updates. For a more detailed description of the meta-RL problem setting, we refer the reader to~\cite{rothfuss2019promp}.

We seek to verify the ability of MMAML to learn to adapt to tasks sampled from multimodal task distributions based on limited interaction with the environment. We do so by evaluating MMAML and the baselines on four continuous control environments using the MuJoCo physics simulator~\cite{todorov2012mujoco}. In all of the environments, the agent is rewarded on every time step for minimizing the distance to the goal. The goals are sampled from multimodal goal distributions with environment specific parameters. The agent does not observe the location of the goal directly but has to learn to find it based on the reward instead. To provide intuition on the environments, illustrations of the robots are presented in \myfig{fig:rl_example}. Examples of trajectories are presented in \myfig{fig:rl_trajectories} for \textsc{Point Mass} and in \myfig{fig:rl_screenshots} for \textsc{Ant} and \textsc{Reacher}. Complete details of the environments and goal distributions can be found in the supplementary material.

\noindent{\bf Baselines and Our Approach.} To identify the mode of a task distribution with MMAML, we run the initial policy to interact with the environment and collect a batch of trajectories. These initial trajectories are used for two purposes: computing the adapted parameters using a gradient-based update and modulating the updated parameters based on the task embedding $\upsilon$ computed by the modulation network. The modulation vectors $\tau$ are kept fixed for the subsequent gradient updates. Descriptions of the network architectures and training hyperparameters are deferred to the supplementary material. Due to credit-assignment problems present in the MAML for RL algorithm~\cite{finn_model-agnostic_2017} as identified in \cite{rothfuss2019promp}, we optimize our policies and modulation networks with ProMP~\cite{rothfuss2019promp} algorithm, which resolves these issues.

We use ProMP both as the training algorithm for MMAML and as a baseline. Multi-ProMP is an artificial baseline to show the performance of training one policy for each mode using ProMP. In practice we train an agent for only one of the modes since the task distributions are symmetric and the agent is initialized to a random pose.

\noindent{\bf Results.} The results for the meta-RL experiments presented in \mytable{table:rl} show that MMAML consistently outperforms the unmodulated ProMP. The good performance of Multi-ProMP, which only considers a single mode suggests that the difficulty of adaptation in our environments results mainly from the multiple modes. We find that the difficulty of the RL tasks does not scale directly with the number of modes, \ie the performance gap between MMAML and ProMP for \textsc{Point Mass} with 6 modes is smaller than the gap between them for 4 modes. We hypothesize the more distinct the different modes of the task distribution are, the more difficult it is for a single policy initialization to master. Therefore, adding intermediate modes (going from 4 to 6 modes) can in some cases make the task distribution easier to learn.

The tSNE visualizations of embeddings of random tasks in the \textsc{Point Mass} and \textsc{Reacher} domains are presented in \myfig{fig:tSNE}. The clearly clustered embedding space shows that the task encoder is capable of identifying the task mode and the good results MMAML achieves suggest that the modulation network effectively utilizes the task embeddings to tackle the multimodal task distribution.


\Skip{
\begin{enumerate}
    \item \sun{talk about few-shot RL problem}
    \item \sun{talk about what we want to examine + the high level idea of environment design + brief descriptions of envs}
    \item \sun{results}
    \item \sun{short conclusion of RL exp}
\end{enumerate}
}

\input{text/table/reinforcement_learning_table.tex}

%% file: text/table/reinforcement_learning_table.tex
\begin{table*}[t]
    \centering
    \small
    \caption{\small The mean and standard deviation of cumulative reward per episode for multimodal reinforcement learning problems with 2, 4 and 6 modes reported across 3 random seeds. Multi-ProMP is ProMP trained on an easier task distribution which consists of a single mode of the multimodal distribution to provide an approximate upper limit on the performance on each task.}
    \scalebox{0.85}{

    \begin{tabular}{ccccccccc} 
    \toprule
    \multirow{2}{*}{\textbf{Method}} &
    \multicolumn{3}{c}{\textbf{\textsc{Point Mass 2D}}} &
    \multicolumn{3}{c}{\textbf{\textsc{Reacher}}} &
    \multicolumn{2}{c}{\textbf{\textsc{Ant}}}
    \\ 
    \cmidrule(lr){2-4}
    \cmidrule(lr){5-7}
    \cmidrule(lr){8-9}
    
    & 
    \textbf{2 Modes} &
    \textbf{4 Modes} &
    \textbf{6 Modes} & 
    \textbf{2 Modes} &
    \textbf{4 Modes} &
    \textbf{6 Modes} &
    \textbf{2 Modes} &
    \textbf{4 Modes}
    \\
    \midrule
    ProMP~\citep{rothfuss2019promp} & 
     -397 $\pm$ 20 &
     -523 $\pm$ 51 &
     -330 $\pm$ 10&
     -12 $\pm$ 2.0 &
     -13.8 $\pm$ 2.5 &
     -14.9 $\pm$ 2.9  &
    -761 $\pm$ 48 &
    -953 $\pm$ 46
     \\ 
    Multi-ProMP & 
     -109 $\pm$ 6 & 
     -109 $\pm$ 6 &
     -92 $\pm$ 4 &
    -4.3 $\pm$ 0.1 &
    -4.3 $\pm$ 0.1 &
    -4.3 $\pm$ 0.1 &
    -624 $\pm$ 38 &
    -611 $\pm$ 31
     \\ 
    \midrule
    Ours & 
    -136 $\pm$ 8 &
    -209 $\pm$ 32 &
    -169 $\pm$ 48 &
    -10.0 $\pm$ 1.0 &
    -11.0 $\pm$ 0.8 &
    -10.9 $\pm$ 1.1 &
    -711 $\pm$ 25 &
    -904 $\pm$ 37
    \\
    \bottomrule
    \end{tabular}}
    \label{table:rl}
\end{table*}

%% file: text/conclusion.tex
\section{Conclusion}
\label{sec:conclusion}
We present a novel approach that is able to leverage the strengths of both model-based and model-agnostic meta-learners to discover and exploit the structure of multimodal task distributions. 
Given a few samples from a target task, our modulation network first identifies the mode of the task distribution and then modulates the meta-learned prior in a parameter space. Next, the gradient-based meta-learner efficiently adapts to the target task through gradient updates. 
We empirically observe that our modulation network is capable of effectively recognizing the task modes and producing embeddings that captures the structure of a multimodal task distribution.
We evaluated our proposed model in multimodal few-shot regression, image classification and reinforcement learning, and achieved superior generalization performance on tasks sampled from multimodal task distributions.

%% file: text/acknowledgment.tex
\section*{Acknowledgment} 

This work was initiated when Risto Vuorio was at SK T-Brain and was partially supported by SK T-Brain. 
The authors are grateful for the fruitful discussion with Kuan-Chieh Wang, Max Smith, and Youngwoon Lee.
The authors appreciate the anonymous NeurIPS reviewers as well as the anonymous reviewers who rejected this paper but provided constructive feedback for improving this paper in previous submission cycles.

%% file: text/appendix.tex
\appendix

\section{Details on Modulation Operators}
\label{sec:method_details}

\textbf{Attention based modulation} has been widely used in modern deep learning models and has proved its effectiveness across various tasks~\citep{yang2016stacked,mnih2014recurrent,zhang_self-attention_2018,xu2015show}. Inspired by the previous works, we employed attention to modulate the prior model. In concrete terms, attention over the outputs of all neurons (Softmax) or a binary gating value (Sigmoid) on each neuron's output is computed by the modulation network. These modulation vectors $\tau$ are then used to scale the pre-activation of each neural network layer $\mathbf{F}_{\theta}$, such that $\mathbf{F}_{\phi} = \mathbf{F}_{\theta} \otimes \tau$. Note that here $\otimes$ represents a channel-wise multiplication. 
 
\textbf{Feature-wise linear modulation (FiLM)} has been proposed to modulate neural networks for achieving the conditioning effects of data from different modalities. We adopt FiLM as an option for modulating our task network parameters. Specifically, the modulation vectors $\tau$ are divided into two components $\tau_{\gamma}$ and $\tau_{\beta}$ such that for a certain layer of the neural network with its pre-activation $\mathbf{F}_{\theta}$, we would have $\mathbf{F}_{\phi} = \mathbf{F}_{\theta} \otimes \tau_{\gamma} + \tau_{\beta}$. It can be viewed as a more generic form of attention mechanism. Please refer to ~\cite{perez_film:_2017} for the complete details. In a recent few-shot image classification paper \citep{oreshkin_tadam:_2018}, FiLM modulation is used in a metric learning model and achieves high performance. Similarly, employing FiLM modulation has been shown effective on a variety of tasks such as image synthesis~\cite{karras2019style, park2019semantic, huh2019feedback, almahairi2018augmented}, visual question answering~\cite{perez_film:_2017, perez2017learning}, style transfer~\cite{dumoulin2017learned}, recognition~\cite{hu2018squeeze, xie2018attentional}, reading comprehension~\cite{dhingra2016gated}, etc.

\section{Further Discussion on Related Works}

\paragraph{Discussions on Task-Specific Adaptation/Modulation.}
As mentioned in the related work of the main text, some recent works~\cite{oreshkin_tadam:_2018,YeHZS2018Learning,lee2018gradient} leverage the task-specific adaptation or modulation to achieve few-shot image classification. 
Now we discuss about them in details. 
\cite{oreshkin_tadam:_2018} propose to learn a task-specific network that adapts the weight of the visual embedding networks via feature-wise linear modulation (FiLM)~\cite{perez_film:_2017}. 
Similarly, ~\cite{YeHZS2018Learning} learns to perform similar task-specific adaptation for few-shot image classification via Transformer~\cite{vaswani_attention_2017}.
\cite{lee2018gradient} learns a visual embedding network with a task-specific metric and task-agnostic parameters, where the task-specific metric can be update via a fixed steps of gradient updates similar to ~\cite{finn_meta-learning_2017}.
In contrast, we aim to leverage the power of task-specific modulation to develop a more powerful model-agnostic meta-learning framework, 
which is able to effectively adapt to tasks sampled from a multimodal task distribution.
Note that our proposed framework is capable of solving few-shot regression, classification, and reinforcement learning tasks.

\section{Baselines}
Since we aim to develop a general model-agnostic meta-learning framework, the comparison to methods that achieved great performance on only an individual domain are omitted.

\paragraph{Image Classification.}
While Prototypical networks~\cite{snell2017prototypical}, Proto-MAML~\cite{metadataset}, and TADAM~\cite{oreshkin_tadam:_2018} learn a metric space for comparing samples and therefore are not directly applicable to regression and reinforcement learning domains, we believe it would be informative to evaluate those methods on our multimodal image classification setting. For this purpose, we refer the readers to a recent work~\cite{metadataset} which presents extensive experiments on a similar multimodal setting with a wide range of methods, including model-based (RNN-based) methods, model-agnostic meta-learners, and metric-based methods.

\paragraph{Reinforcement Learning.}
We believe comparing MMAML to ProMP~\cite{rothfuss2019promp} on reinforcement learning tasks highlights the advantage of using a separate modulation network in addition to the task network, given that in the reinforcement learning setting MMAML uses ProMP as the optimization algorithm. Besides ProMP, Bayesian MAML~\cite{kim_bayesian_2018} presents an appealing baseline for multimodal task distributions. We tried to run Bayesian MAML on our multimodal task distributions but had technical difficulties with it. The source code for Bayesian MAML in classification and regression is not publicly available.

\Skip{
\paragraph{Discussions on Probabilistic/Bayesian MAML}
~\citet{finn_probabilistic_2018, kim_bayesian_2018} propose extensions of MAML, which samples a model for each new task from a model distribution, allowing the model to deal with a wider range of tasks. We believe the model distribution is still unimodal (with a Gaussian prior), which is not well-designed to address multimodal task distributions (similar to MAML).

\risto{Removed this 2019-07-27 because we don't have evidence to back the claim.}
}

\section{Additional Experimental Details}
\label{sec:experimental_details}
\subsection{Regression}

\begin{figure}[t]
    \centering
    \begin{tabular}{ccc}
    \includegraphics[width=.3\textwidth]{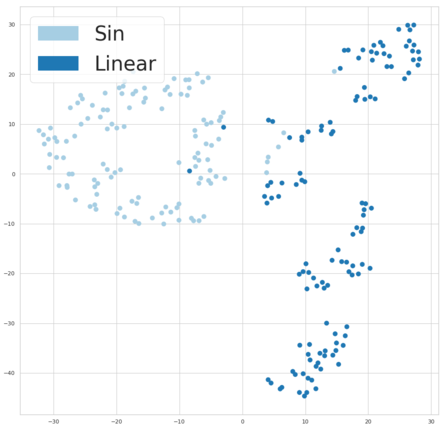} &
    \includegraphics[width=.3\textwidth]{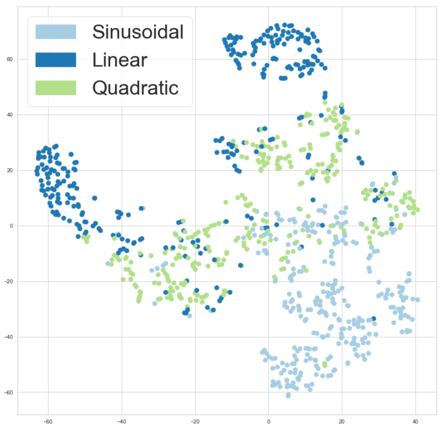} &
    \includegraphics[width=.3\textwidth]{figures/regression_tSNE_5modes.png} \\
    \small{(a) 2-Mode Regression} & (b) \small{3-Mode Regression} & 
    (c) \small{5-Mode Regression}
    \end{tabular}
    
    \caption{ \small
        tSNE plots of the task embeddings produced by our model from randomly sampled tasks for regression. We choose to visualize the corresponding task embeddings of two modes, three modes and five modes.
    }
    \label{fig:regression_tSNE}
\end{figure}

\subsubsection{Setups} 
To form multimodal task distributions for regression, we consider a family of functions including
sinusoidal functions (in forms of $A \cdot \sin{w \cdot x + b} + \epsilon$, with $A \in [0.1, 5.0]$, $w \in [0.5, 2.0]$ and $b \in [0, 2\pi ]$), 
linear functions (in forms of $A \cdot x + b$, with $A \in [-3, 3]$ and $b \in [-3, 3] $), quadratic functions (in forms of $A \cdot (x - c)^2 + b$, with $A \in [-0.15, -0.02] \cup [0.02, 0.15]$, $c \in [-3.0, 3.0]$ and $b \in [-3.0, 3.0]$ ), $\ell_1$ norm function (in forms of $A \cdot |x - c| + b$, with $A \in [-0.15, -0.02] \cup [0.02, 0.15]$, $c \in [-3.0, 3.0]$ and $b \in [-3.0, 3.0]$), and hyperbolic tangent function (in forms of $A \cdot tanh(x - c) + b$, with $A \in [-3.0, 3.0]$, $c \in [-3.0, 3.0]$ and $b \in [-3.0, 3.0]$). Gaussian observation noise with $\mu=0$ and $\epsilon=0.3$ is added to each data point sampled from the target task. In all the experiments, $K$ is set to $5$ and $L$ is set to $10$. We report the mean squared error (MSE) as the evaluation criterion. 
Due to the multimodality and uncertainty, this setting is more challenging comparing to~\citep{finn_model-agnostic_2017}. 

\subsubsection{Models and Optimization} 
In the regression task, we trained a 4-layer fully connected neural network with the hidden dimensions of $100$ and ReLU non-linearity for each layer, as the base model for both MAML and MMAML. In MMAML, an additional model with a Bidirectional LSTM of hidden size $40$ is trained to generate $\tau$ and to modulate each layer of the base model. We used the same hyper-parameter settings as the regression experiments presented in ~\cite{finn_model-agnostic_2017} and used Adam~\cite{kingma2014adam} as the meta-optimizer. For all our models, we train on 5 meta-train examples and evaluate on 10 meta-val examples to compute the loss. 

\subsubsection{Evaluation Protocol} 
In the evaluation of regression experiments, we samples 25,000 tasks for each task mode and evaluate all models with 5 gradient steps during the adaptation (if applicable), with the adaptation learning rate set to be the one models learned with. Therefore, the results for 2 mode experiments is computed over 50,000 tasks, corresponding 3 mode experiment is computed over 75,000 tasks and 5 mode has 125,000 tasks in total. We evaluate all methods over the function range between -5 and 5, and report the accumulated mean squared error as performance measures.

\subsubsection{Effect of Modulation and Adaptation}

\begin{figure}[thb]
\begin{tabular}{cc}
    \tabcolsep 5pt
    \aligntop{
        \begin{minipage}{.45 \textwidth}
        	\centering
        	\tabcolsep 8pt
        	\small
            \caption{
            	\small{5-mode Regression: Visualization with Linear \& Quadratic Function.} \vspace{-1em} \label{fig:eff_reg}
        	}
        	\scalebox{0.9}{
        	\begin{tabular}{cc}
        	    \scriptsize{\textbf{Linear}} & \scriptsize{\textbf{Quadratic}} \\
        		\includegraphics[width=0.475\textwidth,trim={0.8cm 0 0.2cm 0},clip]{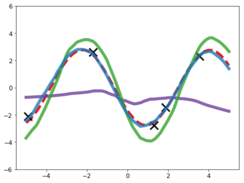} &
        		\includegraphics[width=0.475\textwidth,trim={0.8cm 0 0.2cm 0},clip]{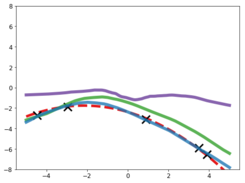} \\
                \multicolumn{2}{c}{\includegraphics[width=0.98\textwidth]{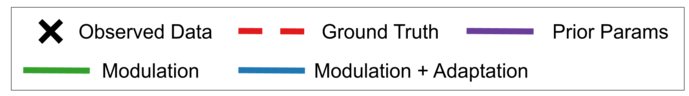}}
        	\end{tabular}
        	}
    	\end{minipage} 
	} &
    \aligntop{
        \begin{minipage}{.45\textwidth}
            \centering
            \small
            \tabcolsep 20pt
            \captionof{table}{\small 5-mode Regression: Performance measured in mean squared error (MSE).}
            \scalebox{1.0}{
            \begin{tabular}{lr} 
            \toprule
            \textbf{MMAML} & \textbf{MSE} \\
            \midrule
            {Prior Params}     & {17.299} \\
            + \textbf{Modulation} & {2.166}  \\
            + \textbf{Adaptation} & {0.868} \\
            \bottomrule
            \end{tabular}}
            \label{table:eff_reg}
        \end{minipage}
    }
\end{tabular}
\end{figure}

We analyze the effect of modulation and adaptation steps on the regression experiments.
Specifically, we show both the qualitative and quantitative results on the 5-mode regression task, and plot the induced function curves as well as measure the Mean Squared Error (MSE) after applying modulation step or both modulation and adaptation step.
Note that MMAML starts from a learned prior parameters (denoted as \textit{prior params}), and then sequentially performs modulation and adaptation steps. The results are shown in the Figure~\ref{fig:eff_reg} and Table~\ref{table:eff_reg}. We see that while inference with prior parameters itself induces high error, adding modulation as well as further adaptation can significantly reduce such error. We can see that the modulation step is trying to seek a rough solution that captures the shape of the target curve, and the gradient based adaptation step refines the induced curve.

\subsection{Image Classification}
\subsubsection{Meta-dataset}

\begin{figure}[ht]
	\centering
	\small
	\includegraphics[width=0.95\textwidth]{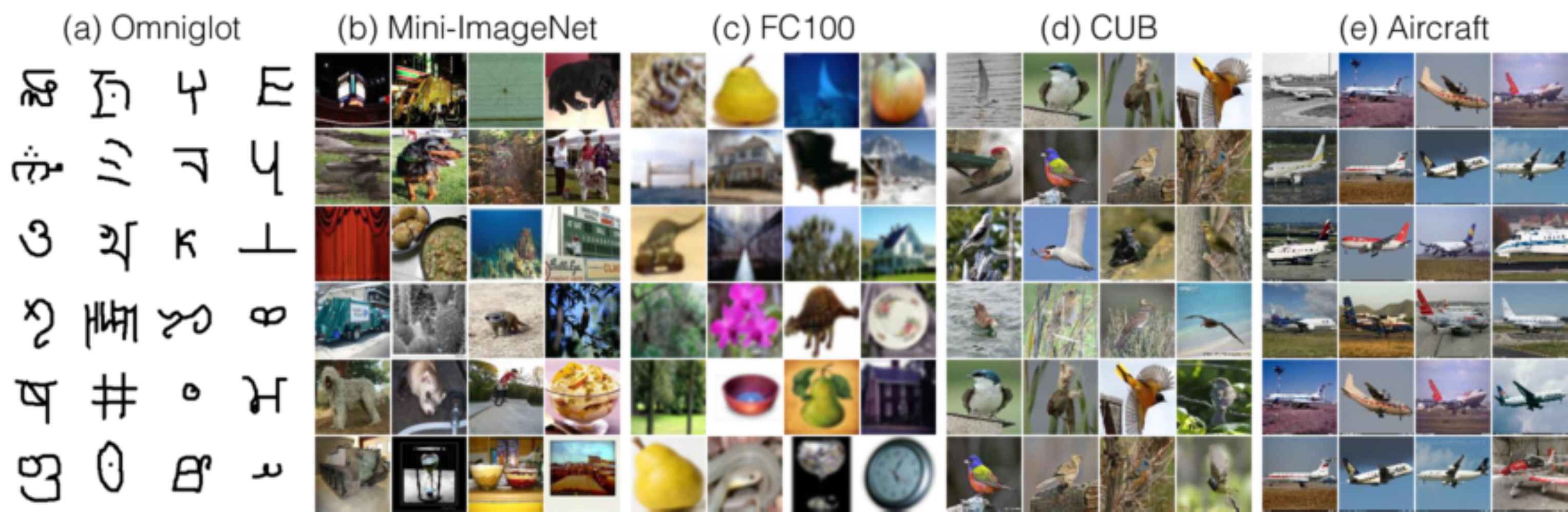}
	\caption{
    	\small Examples of images from all the datasets.
	}
	\label{fig:summary}
\end{figure}

To create a meta-dataset by merging multiple datasets, we utilize five popular datasets: \textsc{Omniglot}, \textsc{Mini-ImageNet}, \textsc{FC100}, \textsc{CUB}, and \textsc{Aircraft}.
The detailed information of all the datasets are summarized in \mytable{table:classification_dataset}.
To fit the images from all the datasets to a model, we resize all the images to $84 \times 84$.
The images randomly sampled from all the datasets are shown in \myfig{fig:summary}, demonstrating a diverse set of modes.


\input{text/table/appendix_classification_dataset.tex}

\input{text/table/appendix_classification_hyperparameter.tex}

\input{text/table/appendix_classification.tex}

\subsubsection{Hyperparameters}
We present the hyperparameters for all the experiments in \mytable{table:classification_hyperparameter}.
We use the same set of hyperparameters to train our model and MAML for all experiments, 
except that we use a smaller meta batch-size for 20-way tasks and train the jobs for more iterations due to the limited memory of GPUs that we have access to. 

We use 15 examples per class for evaluating the post-update meta-gradient for all the experiments, following~\cite{finn_model-agnostic_2017, ravi_optimization_2017}. All the trainings use the Adam optimizer~\cite{kingma2014adam} with default hyperparameters.

For Multi-MAML, since we train a MAML model for each dataset,
it gives us the freedom to use different sets of hyperparameters for different datasets
We tried our best to find the best hyperparameters for each dataset.


\subsubsection{Network Architectures}

\paragraph{Task Network.}
For the task network, we use the exactly same architecture as the MAML convolutional network proposed in~\cite{finn_model-agnostic_2017}. 
It consists of four convolutional layers with the channel size $32$, $64$, $128$, and $256$, respectively. All the convolutional layers have a kernel size of $3$ and stride of $2$. A batch normalization layer follows each convolutional layer, followed by ReLU.
With the input tensor size of $(n\cdot k)\times84\times84\times3$ for a $n$-way $k$-shot task, the output feature maps after the final convolutional layer have a size of $(n\cdot k)\times6\times6\times256$.
The feature maps are then average pooled along spatial dimensions, resulting feature vectors with a size of $(n\cdot k)\times256$.
A linear fully-connected layer takes the feature vector as input, and produce a classification prediction with a size of $n$ for n-way classification tasks.

\paragraph{Task Encoder.}
For the task encoder, we use the exactly same architecture as the task network. It consists of four convolutional layers with the channel size $32$, $64$, $128$, and $256$, respectively. All the convolutional layers have a kernel size of $3$, stride of $2$, and use valid padding. A batch normalization layer follows each convolutional layer, followed by ReLU.
With the input tensor size of $(n\cdot k)\times84\times84\times3$ for a $n$-way $k$-shot task, the output feature maps after the final convolutional layer have a size of $(n\cdot k)\times6\times6\times256$.
The feature maps are then average pooled along spatial dimensions, resulting feature vectors with a size of $(n\cdot k)\times256$.
To produce an aggregated embedding vector from all the feature vectors representing all samples, we perform an average pooling, resulting a feature vector with a size of $256$.
Finally, a fully-connected layer followed by ReLU takes the feature vector as input, and produce a task embedding vector $\upsilon$ with a size of $128$.

\paragraph{Modulation MLPs}. 
Since the task network consists of four convolutional layers with the channel size $32$, $64$, $128$, and $256$ 
and modulating each of them requires producing both $\tau_\gamma$ and $\tau_\beta$,
we employ four linear fully-connected layers to convert the task embedding vector $\upsilon$ to 
$\{\tau_{\gamma_1}$, $\tau_{\beta_1}\}$ (with a dimension of $32$),
$\{\tau_{\gamma_2}$, $\tau_{\beta_2}\}$ (with a dimension of $64$),
$\{\tau_{\gamma_3}$, $\tau_{\beta_3}\}$ (with a dimension of $128$), and
$\{\tau_{\gamma_4}$, $\tau_{\beta_4}\}$ (with a dimension of $256$).
Note the modulation for each layer is performed by $\theta_i \odot \gamma_i + \beta_i$, 
where $\odot$ denotes the Hadamard product.

\begin{figure*}[t]
    \centering
    \begin{tabular}{ccc}
    \includegraphics[width=0.3\textwidth]{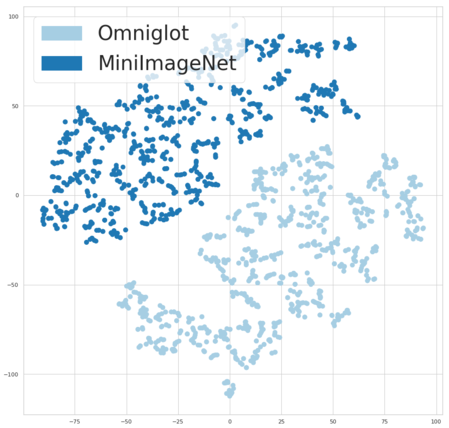} &
    \includegraphics[width=0.3\textwidth]{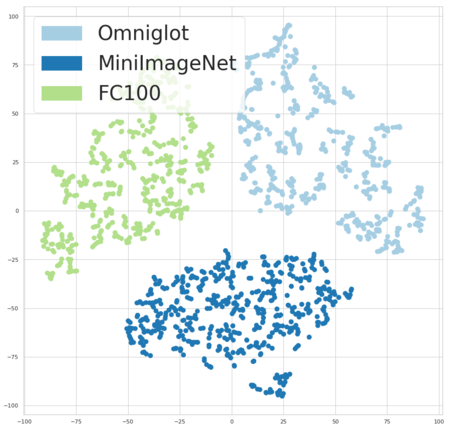} &    
    \includegraphics[width=0.3\textwidth]{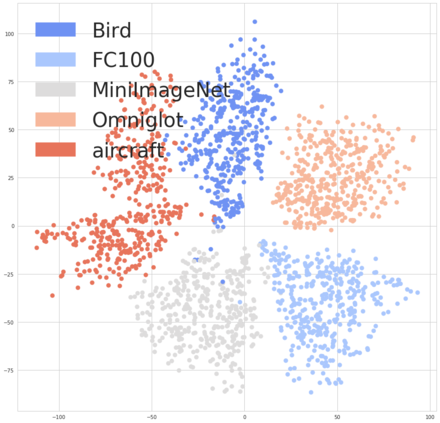} \\
    \small{(a) 2-mode classification} &
    \small{(b) 3-mode classification} &     
    \small{(c) 5-mode classification}
    \end{tabular}
    \caption{ \small
        tSNE plots of task embeddings produced in multimodal few-shot image classification domain. 
        (a) 2-mode 5-way 1-shot (b) 3-mode 5-way 1-shot (c) 5-mode 5-way 5-shot.
        \label{fig:appendix_classification_tSNE}
    }
\end{figure*}

\subsection{Reinforcement Learning}
\label{subsec:rl_appendix}

\begin{figure*}[t]
    \centering
    \begin{tabular}{ccc}
    \includegraphics[width=0.3\textwidth]{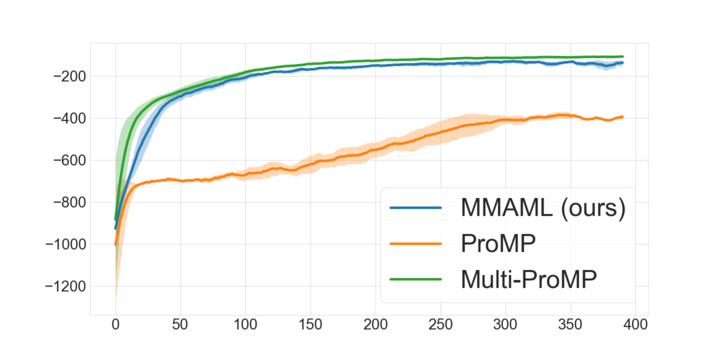} &
    \includegraphics[width=0.3\textwidth]{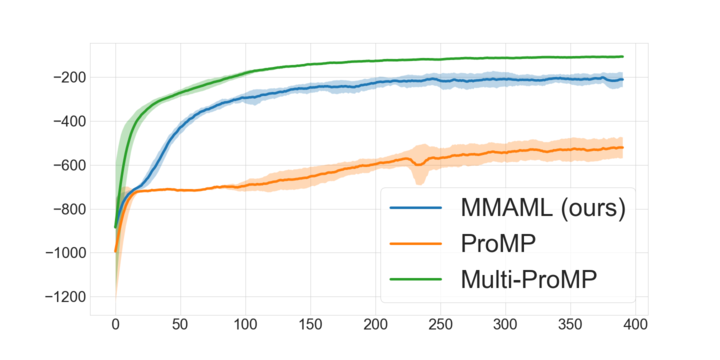} &
    \includegraphics[width=0.3\textwidth]{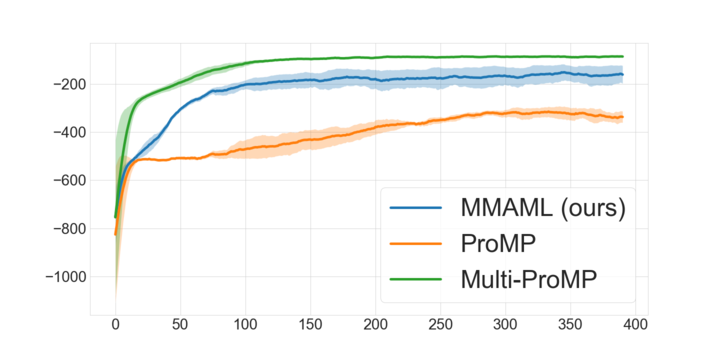} \\
    \small{(a) \textsc{Point Mass} 2 Modes} & \small{(b) \textsc{Point Mass} 4 Modes} & \small{(c) \textsc{Point Mass} 6 Modes} \\
    
    \includegraphics[width=0.3\textwidth]{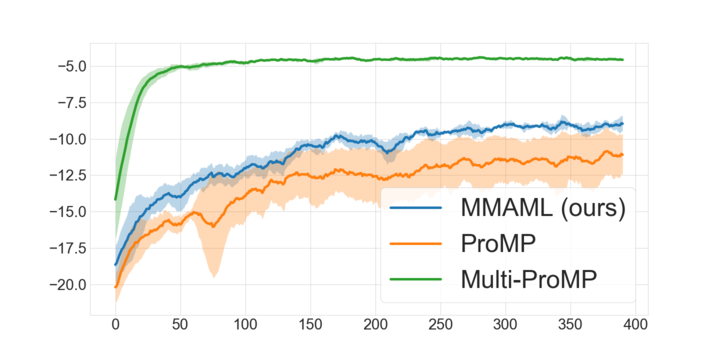} &
    \includegraphics[width=0.3\textwidth]{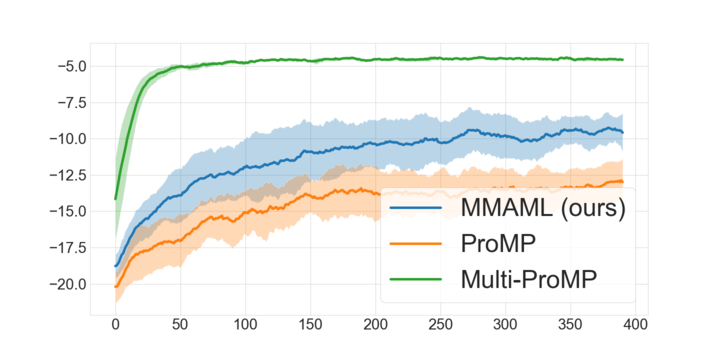} &
    \includegraphics[width=0.3\textwidth]{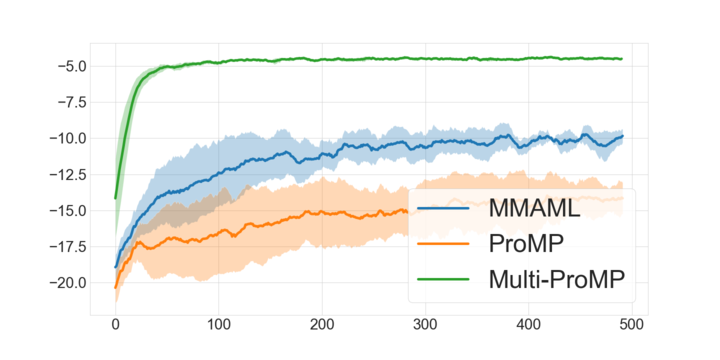} \\
    \small{(a) \textsc{Reacher} 2 Modes} & \small{(b) \textsc{Reacher} 4 Modes} & \small{(c) \textsc{Reacher} 6 Modes} \\
    
    \includegraphics[width=0.3\textwidth]{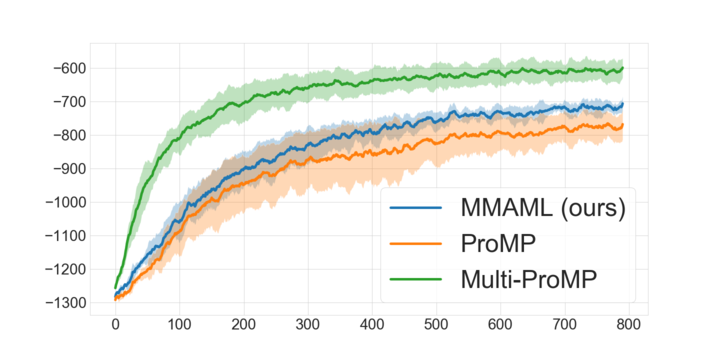} & \includegraphics[width=0.3\textwidth]{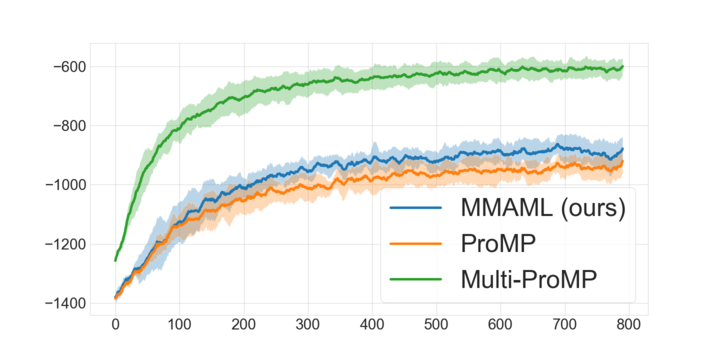} & \\
    \small{(a) \textsc{Ant} 2 Modes} &  \small{(a) \textsc{Ant} 4 Modes} &
    
    \end{tabular}
    \caption{
    \small
    Training curves for MMAML and ProMP in reinforcement learning environments. The curves indicate the average return per episode after gradient-based updates and modulation. The shaded region indicates standard deviation across three random seeds. The curves have been smoothed by averaging the values within a window of 10 steps.
    }
    \label{fig:rl_training_curves}
\end{figure*}

\subsubsection{Environments}

The training curves for all environments are presented in \myfig{fig:rl_training_curves}.

\paragraph{\textsc{Point Mass}}. We consider three variants of the \textsc{Point Mass} environment with 2, 4, and 6 modes. The agent controls a point mass by outputting changes to the velocity. At every time step the agent receives the negative euclidean distance to the goal as the reward. The goals are sampled from a multimodal goal distribution by first selecting the mode center and then adding Gaussian noise to the goal location. In the 4 mode variant the modes are the points $(-5, -5)$, $(-5, 5)$, $(5, -5)$, $(5, 5)$. In the 2 mode variant the modes are the points $(-5, -5)$, $(5, 5)$. In the 6 mode variant the modes are the vertices of a regular hexagon with at distance $5$ from the origin. All variants have noise scale of $2.0$. Visualizations of agent trajectories can be found in~\myfig{fig:rl_extra_results}.

\paragraph{\textsc{Reacher}}. We consider three variants of the \textsc{Reacher} environment with 2, 4, and 6 modes. The agent controls a 2-dimensional robot arm with three links simulated in the MuJoCo \cite{todorov2012mujoco} simulator. The goal distribution is similar to the goal distributions in \textsc{Point Mass} but different parameters are used to match the scale of the environment. The reward for the environment is 
$$R(s, a) = - 1 * (x_{point} - x_{goal})^2 - \|a\|^2$$
where $x_{point}$ is the location of the point of the arm, $x_{goal}$ if the location of the goal and $a$ is the action chosen by the agent. The modes of the goal distribution in the 4 mode variant are located at $(-0.225, -0.225)$, $(0.225, -0.225)$, $(-0.225, 0.225)$, $(0.225, 0.225)$ and the goal noise has scale of $0.1$. In the 2 mode variant the modes are located at $(-0.225, -0.225)$, $(0.225, 0.225)$ and the noise scale is $0.1$. In the 6 mode variant the mode centers are the vertices of a regular hexagon with distance to the origin of $0.318$ and the noise scale is $0.1$.

\paragraph{\textsc{Ant}}. We consider two variants of the \textsc{Ant} environment with two and four modes. The agent controls an ant robot with four limbs simulated in the MuJoCo \cite{todorov2012mujoco} simulator.
The reward for the environment is 
$$R(s, a) = - 1 * (x_{torso} - x_{goal})^2 - \lambda_{control} * \|a\|^2$$
where $x_{torso}$ is the location of the torso of the robot, $x_{goal}$ if the location of the goal, $\lambda_{control} = 0.1$ is the weighting for the control cost and $a$ is the action chosen by the agent.
The modes of the goal distribution in the 4 mode variant are located at $(-4, 0)$, $(-2, 3.46)$, $(2, 3.46)$, $(4.0, 0)$ and the goal noise has scale of $0.8$. In the 2 mode variant the modes are located at $(-4.0, 0)$, $(4.0, 0)$ and the noise scale is $0.8$.


\subsubsection{Network Architectures and Hyperparameters}
For all RL experiments we use a policy network with two 64-unit hidden layers. The modulation network in RL tasks consists of a GRU-cell and post processing layers. The inputs to the GRU are the concatenated observations, actions and reward for each trajectory. The trajectories are processed separately. An MLP is used to process the last hidden states of each trajectory. The outputs of the MLPs are averaged and used by another MLP to compute the modulation vectors $\tau$. All MLPs have a single hidden layer of size 64.

We sample 40 tasks for each update step. For each gradient step for each task we sample 20 trajectories. The hyperparameters, which differ from setting to setting are presented in \mytable{table:rl_hyperparameters}.

\input{text/table/appendix_reinforcement_learning_hyperparameter_table.tex}

\section{Additional Experimental Results}
\label{sec:additional_results}
\subsection{Regression}

We show visualization of embeddings for regression experiments with a varying number of task modes as \myfig{fig:regression_tSNE}. We observe a linear separation in the two task modes and three task modes scenarios, which indicates that our method is capable of identifying data from different task modes. On the visualization of five task mode, we observe that data from linear, transformed $\ell_1$ norm and hyperbolic tangent functions cluttered. This is due to the fact that those functions are very similar to each other, especially with the Gaussian noise we added in the output space.



    

\begin{figure}[ht]
	\centering
	\small
	\tabcolsep 1pt
	\begin{tabular}{ccccc}
		\multicolumn{5}{c}{\includegraphics[width=0.75\textwidth]{figures/regression/regression_legend}} \\
	    \scriptsize{\textbf{Sinusoidal}} & \scriptsize{\textbf{Linear}} &
	    \scriptsize{\textbf{Quadratic}}  & \scriptsize{\textbf{Transformed $\ell_1$ Norm}} & 
	    \scriptsize{\textbf{Tanh}} \\

		\includegraphics[width=0.185\textwidth,trim={1.4cm 0 0.6cm 0},clip]{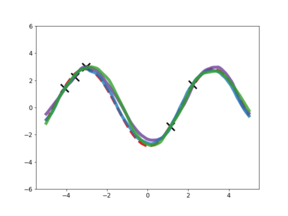} &
		\includegraphics[width=0.185\textwidth,trim={1.4cm 0 0.6cm 0},clip]{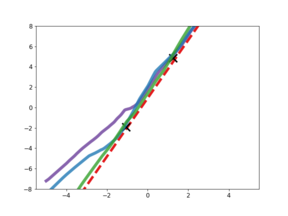} &
		\includegraphics[width=0.185\textwidth,trim={1.4cm 0 0.6cm 0},clip]{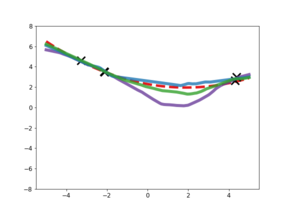} &
		\includegraphics[width=0.185\textwidth,trim={1.4cm 0 0.6cm 0},clip]{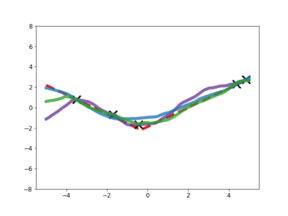} &
		\includegraphics[width=0.185\textwidth,trim={1.4cm 0 0.6cm 0},clip]{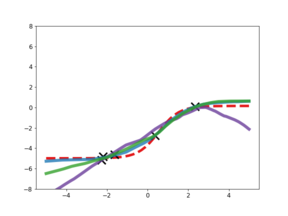} \\
		\includegraphics[width=0.185\textwidth,trim={1.4cm 0 0.6cm 0},clip]{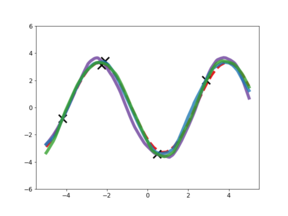} &
		\includegraphics[width=0.185\textwidth,trim={1.4cm 0 0.6cm 0},clip]{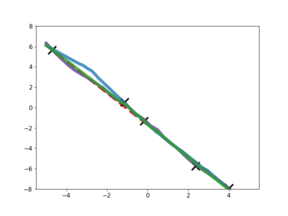} &
		\includegraphics[width=0.185\textwidth,trim={1.4cm 0 0.6cm 0},clip]{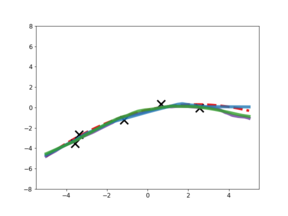} &		\includegraphics[width=0.185\textwidth,trim={1.4cm 0 0.6cm 0},clip]{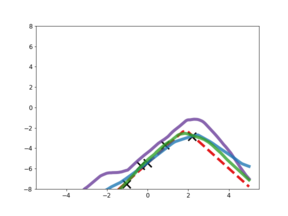} &
		\includegraphics[width=0.185\textwidth,trim={1.4cm 0 0.6cm 0},clip]{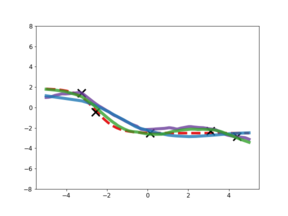} \\
		\includegraphics[width=0.185\textwidth,trim={1.4cm 0 0.6cm 0},clip]{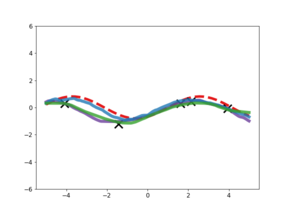} &
		\includegraphics[width=0.185\textwidth,trim={1.4cm 0 0.6cm 0},clip]{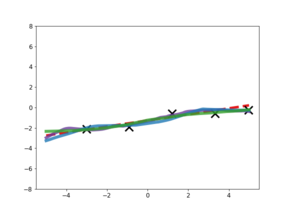} &
		\includegraphics[width=0.185\textwidth,trim={1.4cm 0 0.6cm 0},clip]{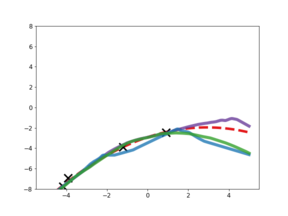} &		\includegraphics[width=0.185\textwidth,trim={1.4cm 0 0.6cm 0},clip]{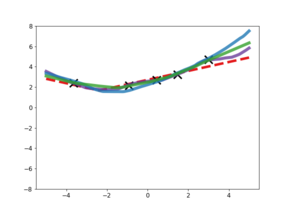} &
		\includegraphics[width=0.185\textwidth,trim={1.4cm 0 0.6cm 0},clip]{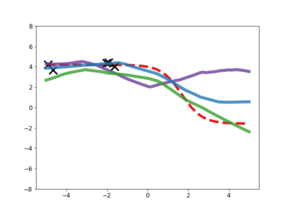} \\
		\includegraphics[width=0.185\textwidth,trim={1.4cm 0 0.6cm 0},clip]{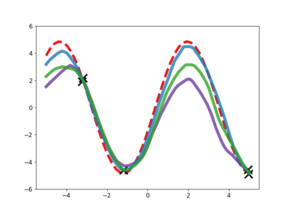} &
		\includegraphics[width=0.185\textwidth,trim={1.4cm 0 0.6cm 0},clip]{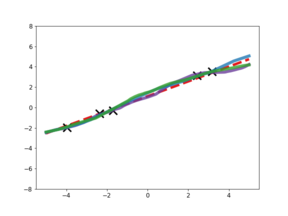} &
		\includegraphics[width=0.185\textwidth,trim={1.4cm 0 0.6cm 0},clip]{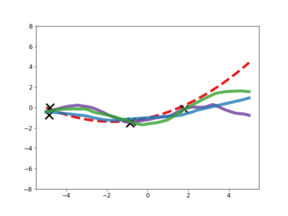} &		\includegraphics[width=0.185\textwidth,trim={1.4cm 0 0.6cm 0},clip]{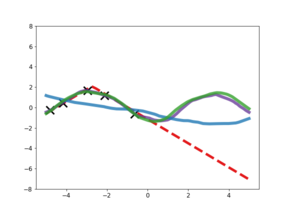} &
		\includegraphics[width=0.185\textwidth,trim={1.4cm 0 0.6cm 0},clip]{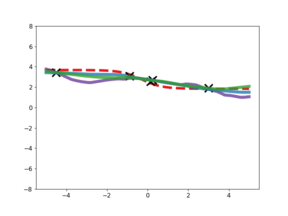} \\
		\includegraphics[width=0.185\textwidth,trim={1.4cm 0 0.6cm 0},clip]{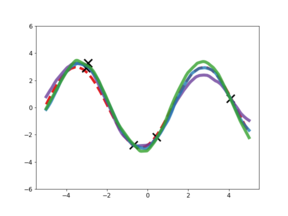} &
		\includegraphics[width=0.185\textwidth,trim={1.4cm 0 0.6cm 0},clip]{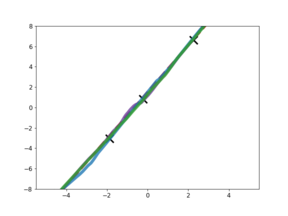} &
		\includegraphics[width=0.185\textwidth,trim={1.4cm 0 0.6cm 0},clip]{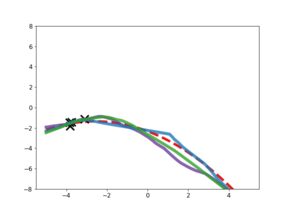} &		\includegraphics[width=0.185\textwidth,trim={1.4cm 0 0.6cm 0},clip]{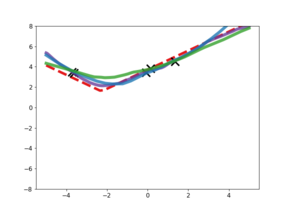} &
		\includegraphics[width=0.185\textwidth,trim={1.4cm 0 0.6cm 0},clip]{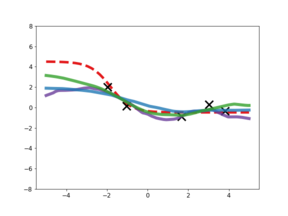} \\
	\end{tabular}
	\caption{
    	\small
	    Additional qualitative results of the regression tasks. MMAML \textbf{after adaptation} vs. other posterior models. 
	}
	\label{fig:regression_B}
\end{figure}

\subsection{Image Classification}
We provide the detailed performance of our method and the baselines on each individual dataset for all 2, 3, and 5 mode experiments, shown in \mytable{table:2mode}, \mytable{table:3mode}, and \mytable{table:5mode}, respectively.
Note that the main paper presents the overall performance (the last columns of each table) on each of 2, 3, and 5 mode experiments.

We found the results on \textsc{Omniglot} and \textsc{Mini-ImageNet} demonstrate similar tendency shown in~\cite{metadataset}.
Note that the performance of \textsc{Omniglot} and \textsc{FC100} might be slightly different from the results reported in the related papers because (1) all the images are resized and tiled along the spatial dimensions, (2) different hyperparamters are used, and (3) different numbers of training iterations.

Additional tSNE plots for predicted task embeddings of 
2-mode 5-way 1-shot classification, 3-mode 5-way 1-shot classification, and 5-mode 20-way 1-shot classification are shown in \myfig{fig:appendix_classification_tSNE}. 

\subsection{Reinforcement Learning}
\begin{figure}
	\tabcolsep 4pt
	\small
    \centering
    \begin{tabular}{cc}
         \includegraphics[width=0.49\textwidth]{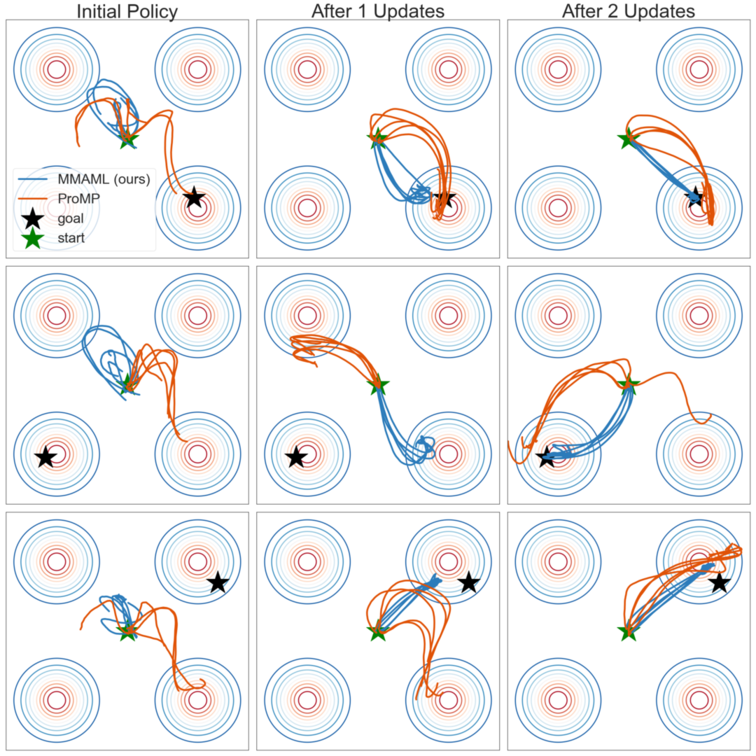} &
         \includegraphics[width=0.49\textwidth]{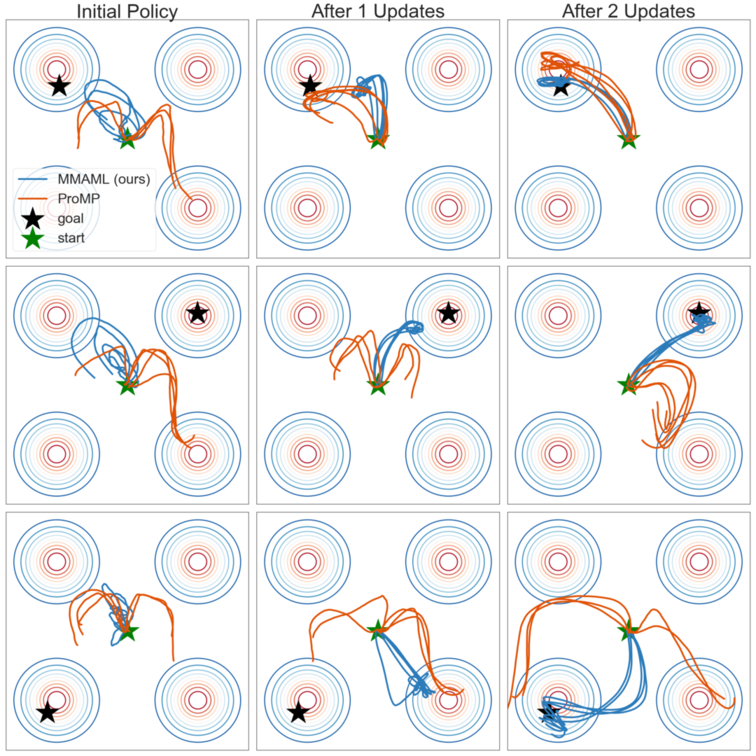} \\
    \end{tabular}
    \caption{Additional trajectories sampled from the point mass environment with MMAML and ProMP for six tasks. The contour plots represents the multimodal task distribution. The stars mark the start and goal locations. The curves depict five trajectories sampled using each method after zero, one and two update steps. In the figure, the modulation step takes place between the initial policy and the step after one update.}
    \label{fig:rl_extra_results}
\end{figure}

Additional trajectories sampled from the 2D navigation environment are presented in \myfig{fig:rl_extra_results}.

%% file: text/table/appendix_classification_dataset.tex
\begin{table*}[t]
    \centering
    \small
    \caption{\small Dataset details.}
    \scalebox{0.8}{
    \begin{tabular}{cccccccc} 
    
    \toprule
    \textbf{Dataset} & 
    \textbf{Train classes} & \textbf{Validation classes} & \textbf{Test classes} & 
    \textbf{Image size} & \textbf{Image channel} & \textbf{Image content} \\
    
    \midrule
    \textsc{Omniglot}      & 4112 & 688 & 1692 & 28 $\times$ 28 & 1 & handwritten characters\\
    \textsc{Mini-ImageNet} & 64   & 16  & 20   & 84 $\times$ 84 & 3 & objects\\
    \textsc{FC100}         & 64   & 16  & 20   & 32 $\times$ 32 & 3 & objects\\
    \textsc{CUB}           & 140  & 30  & 30   & $\sim$ 500 $\times$ 500 & 3 & birds\\
    \textsc{Aircraft}      & 70   & 15  & 15   & $\sim$ 1-2 Mpixels & 3 & aircrafts\\
    
    \bottomrule
    \end{tabular}}
    \label{table:classification_dataset}
\end{table*}

%% file: text/table/appendix_classification_hyperparameter.tex
\begin{table*}[t]
    \centering
    \small
    \caption{\small Hyperparameters for multimodal few-shot image classification experiments.
    We experiment different hyperparameters for each dataset for Multi-MAML. The dataset group \textbf{Grayscale} includes \textsc{Omniglot} and \textbf{RGB} includes \textsc{Mini-ImageNet} and \textsc{FC100}, \textsc{CUB}, and \textsc{Aircraft}.}
    \scalebox{0.75}{
    \begin{tabular}{cccccccc} 
    
    \toprule
    \textbf{Method} & \textbf{Setup} & \textbf{Dataset group} &
    \textbf{Slow lr} & \textbf{Fast lr} & \textbf{Meta bach-size} & 
    \textbf{Number of updates} & \textbf{Training iterations} \\
    
    \midrule
    \multirow{2}{*}{MAML} & 5-way 1-shot
    & \multirow{4}{*}{-} 
    & \multirow{4}{*}{0.001}
    & \multirow{4}{*}{0.05} & \multirow{4}{*}{10} 
    & \multirow{4}{*}{5} & \multirow{4}{*}{60000} \\
    & 5-way 5-shot & \\
    \multirow{2}{*}{MMAML (ours)} & 5-way 1-shot & & & & & \\
    & 5-way 5-shot & \\
    
    \midrule
    \multirow{2}{*}{MAML} & 20-way 1-shot
    & \multirow{4}{*}{-} 
    & \multirow{4}{*}{0.001}
    & \multirow{4}{*}{0.05} & \multirow{4}{*}{5} 
    & \multirow{4}{*}{5} & \multirow{4}{*}{80000} \\
    & 20-way 3-shot & \\
    \multirow{2}{*}{MMAML (ours)} & 20-way 1-shot & & & & & \\
    & 20-way 3-shot & \\    
    
    \midrule
    \multirow{8}{*}{Multi-MAML} & \multirow{2}{*}{5-way 1-shot} & Grayscale 
    & \multirow{8}{*}{0.001} & 0.4 & 10 & 1 & \multirow{4}{*}{60000}\\
     & & RGB & & 0.01 & 4 & 5 \\
     & \multirow{2}{*}{5-way 5-shot} & Grayscale & & 0.4 & 10 & 1 \\
     & & RGB & & 0.01 & 4 & 5\\
     & \multirow{2}{*}{20-way 1-shot} & Grayscale & & 0.1 & 4 & 5 & \multirow{4}{*}{80000} \\
     & & RGB & & 0.01 & 2 & 5\\
     & \multirow{2}{*}{20-way 3-shot} & Grayscale & & 0.1 & 4 & 5 \\
     & & RGB & & 0.01 & 2 & 5\\
    \bottomrule
    
    \end{tabular}
    }
    \label{table:classification_hyperparameter}
\end{table*}

%% file: text/table/appendix_classification.tex
\begin{table*}[t]
    \centering
    \small
    \caption{\small The performance (classification accuracy) on the \textbf{multimodal few-shot image classification} with \textbf{2 modes} on each dataset.}
    \scalebox{0.8}{
    \begin{tabular}{ccccc} 
    
    \toprule
    \multirow{2}{*}{\textbf{Setup}} & \multirow{2}{*}{\textbf{Method}} &
    \multicolumn{2}{c}{\textbf{Datasets}} \\
    & & \textsc{Omniglot} & \textsc{Mini-ImageNet} & \textsc{Overall} \\
    
    \midrule
    \multirow{3}{*}{5-way 1-shot} 
    & MAML          & 89.24\% & 44.36\% & 66.80\% \\
    & Multi-MAML    & 97.78\% & 35.91\% & 66.85\% \\
    & MMAML (ours) & 94.90\% & 44.95\% & 69.93\% \\
   
    \midrule
    \multirow{3}{*}{5-way 5-shot} 
    & MAML          & 96.24\% & 59.35\% & 77.79\% \\
    & Multi-MAML    & 98.48\% & 47.67\% & 73.07\% \\
    & MMAML (ours) & 98.47\% & 59.00\% & 78.73\% \\

    \midrule
    \multirow{3}{*}{20-way 1-shot} 
    & MAML          & 55.36\% & 15.67\% & 35.52\% \\
    & Multi-MAML    & 91.59\% & 14.71\% & 53.15\% \\
    & MMAML (ours) & 83.14\% & 12.47\% & 47.80\% \\

    \bottomrule
    \end{tabular}}
    \label{table:2mode}
\end{table*}

\begin{table*}[t]
    \centering
    \small
    \caption{\small The performance (classification accuracy) on the \textbf{multimodal few-shot image classification} with \textbf{3 modes} on each dataset.}
    \scalebox{0.8}{
    \begin{tabular}{cccccc} 
    
    \toprule
    \multirow{2}{*}{\textbf{Setup}} & \multirow{2}{*}{\textbf{Method}} &
    \multicolumn{3}{c}{\textbf{Datasets}} \\
    & & \textsc{Omniglot} & \textsc{Mini-ImageNet} & \textsc{FC100} & \textsc{Overall} \\
    
    \midrule
    \multirow{3}{*}{5-way 1-shot} 
    & MAML          & 86.76\% & 43.27\% & 33.29\% & 54.55\% \\
    & Multi-MAML    & 97.78\% & 35.91\% & 34.00\% & 55.90\% \\
    & MMAML (ours) & 93.67\% & 41.07\% & 33.67\% & 57.47\% \\
    
    \midrule
    \multirow{3}{*}{5-way 5-shot} 
    & MAML          & 95.11\% & 61.48\% & 47.33\% & 67.97\% \\
    & Multi-MAML    & 98.48\% & 47.67\% & 40.44\% & 62.20\% \\
    & MMAML (ours) & 99.56\% & 60.67\% & 50.22\% & 70.15\% \\
    
    \midrule
    \multirow{3}{*}{20-way 1-shot} 
    & MAML          & 57.87\% & 15.06\% & 11.74\% & 28.22\% \\
    & Multi-MAML    & 91.59\% & 14.71\% & 13.00\% & 39.77\% \\
    & MMAML (ours) & 85.00\% & 13.00\% & 10.81\% & 36.27\% \\

    \bottomrule
    \end{tabular}}
    \label{table:3mode}
\end{table*}

\begin{table*}[t]
    \centering
    \small
    \caption{\small The performance (classification accuracy) on the \textbf{multimodal few-shot image classification} with \textbf{5 modes} on each dataset.}
    \scalebox{0.77}{
    \begin{tabular}{cccccccc} 
    
    \toprule
    \multirow{2}{*}{\textbf{Setup}} & \multirow{2}{*}{\textbf{Method}} &
    \multicolumn{5}{c}{\textbf{Datasets}} \\
    & & \textsc{Omniglot} & \textsc{Mini-ImageNet} & \textsc{FC100} & \textsc{CUB} & \textsc{Aircraft} & \textsc{Overall} \\
    
    \midrule
    \multirow{3}{*}{5-way 1-shot} 
    & MAML          & 83.63\% & 37.78\% & 33.70\% & 86.96\% & 36.74\% & 35.48\% \\
    & Multi-MAML    & 97.78\% & 35.91\% & 34.00\% & 93.44\% & 32.03\% & 27.59\% \\
    & MMAML (ours) & 91.48\% & 42.89\% & 32.59\% & 93.56\% & 38.30\% & 36.82\% \\
    
    \midrule
    \multirow{3}{*}{5-way 5-shot} 
    & MAML          & 89.41\% & 51.26\% & 43.41\% & 82.30\% & 45.80\% & 43.92\% \\
    & Multi-MAML    & 98.48\% & 47.67\% & 40.44\% & 98.56\% & 45.70\% & 47.29\% \\
    & MMAML (ours) & 97.96\% & 51.29\% & 44.08\% & 97.88\% & 53.80\% & 51.53\% \\
    
    \midrule
    \multirow{3}{*}{20-way 1-shot} 
    & MAML          & 59.10\% & 15.49\% & 11.75\% & 59.45\% & 16.31\% & 31.57\% \\
    & Multi-MAML    & 91.59\% & 14.71\% & 13.00\% & 85.46\% & 18.87\% & 30.72\% \\
    & MMAML (ours) & 86.28\% & 14.35\% & 11.59\% & 91.86\% & 24.05\% & 30.89\% \\
    
    
    \bottomrule
    \end{tabular}}
    \label{table:5mode}
\end{table*}

%% file: text/table/appendix_reinforcement_learning_hyperparameter_table.tex
\begin{table*}[t]
    \centering
    \small
    \caption{\small Hyperparameter settings for reinforcement learning.}
    \scalebox{0.78}{
    \begin{tabular}{cccccccc} 
    
    \toprule
    \textbf{Environment} & \textbf{Algorithm} & \textbf{Training Iterations} & \textbf{Trajectory Length} & \textbf{Slow lr} & \textbf{Fast lr} & \textbf{Inner Gradient Steps} & \textbf{Clip eps}\\
    
    \midrule
    \multirow{3}{*}{\textsc{Point Mass}}
    & MMAML
        & \multirow{3}{*}{400}
        & \multirow{3}{*}{100}
        & \multirow{3}{*}{0.0005}
        & \multirow{3}{*}{0.01}
        & \multirow{3}{*}{2}
        & \multirow{3}{*}{0.1} \\
    & ProMP & & & & & & \\
    & Multi-ProMP \\
   
   \midrule
   \multirow{3}{*}{\textsc{Reacher}}
   & MMAML
        & \multirow{3}{*}{800}
        & \multirow{3}{*}{50}
        & \multirow{3}{*}{0.001}
        & \multirow{3}{*}{0.1}
        & \multirow{3}{*}{2}
        & \multirow{3}{*}{0.1} \\
    & ProMP & & & & & & \\
    & Multi-ProMP \\
    
   \midrule
   \multirow{3}{*}{\textsc{Ant}}
    & MMAML
        & \multirow{3}{*}{800}
        & \multirow{3}{*}{250}
        & \multirow{3}{*}{0.001}
        & \multirow{3}{*}{0.1}
        & \multirow{3}{*}{3}
        & \multirow{3}{*}{0.1} \\
    & ProMP & & & & & & \\
    & Multi-ProMP \\
    
    \bottomrule
    \end{tabular}}
    \label{table:rl_hyperparameters}
\end{table*}

%% file: mmaml.bbl
\begin{thebibliography}{10}

\bibitem{almahairi2018augmented}
Amjad Almahairi, Sai Rajeswar, Alessandro Sordoni, Philip Bachman, and Aaron
  Courville.
\newblock Augmented cyclegan: Learning many-to-many mappings from unpaired
  data.
\newblock In {\em International Conference on Machine Learning}, 2018.

\bibitem{andrychowicz_learning_2016}
Marcin Andrychowicz, Misha Denil, Sergio Gomez, Matthew~W. Hoffman, David Pfau,
  Tom Schaul, Brendan Shillingford, and Nando de~Freitas.
\newblock Learning to learn by gradient descent by gradient descent.
\newblock In {\em Advances in Neural Information Processing Systems}, 2016.

\bibitem{bengio1992optimization}
Samy Bengio, Yoshua Bengio, Jocelyn Cloutier, and Jan Gecsei.
\newblock On the optimization of a synaptic learning rule.
\newblock In {\em Preprints Conf. Optimality in Artificial and Biological
  Neural Networks}, 1992.

\bibitem{chen2018a}
Wei-Yu Chen, Yen-Cheng Liu, Zsolt Kira, Yu-Chiang~Frank Wang, and Jia-Bin
  Huang.
\newblock A closer look at few-shot classification.
\newblock In {\em International Conference on Learning Representations}, 2019.

\bibitem{dhingra2016gated}
Bhuwan Dhingra, Hanxiao Liu, Zhilin Yang, William~W Cohen, and Ruslan
  Salakhutdinov.
\newblock Gated-attention readers for text comprehension.
\newblock In {\em Annual Meeting of the Association for Computational
  Linguistics}, 2017.

\bibitem{duan2016rl}
Yan Duan, John Schulman, Xi~Chen, Peter~L Bartlett, Ilya Sutskever, and Pieter
  Abbeel.
\newblock {R}{L} $^ 2$: Fast reinforcement learning via slow reinforcement
  learning.
\newblock {\em arXiv preprint arXiv:1611.02779}, 2016.

\bibitem{dumoulin2017learned}
Vincent Dumoulin, Jonathon Shlens, and Manjunath Kudlur.
\newblock A learned representation for artistic style.
\newblock In {\em International Conference on Learning Representations}, 2017.

\bibitem{finn_model-agnostic_2017}
Chelsea Finn, Pieter Abbeel, and Sergey Levine.
\newblock Model-{Agnostic} {Meta}-{Learning} for {Fast} {Adaptation} of {Deep}
  {Networks}.
\newblock In {\em International Conference on Machine Learning}, 2017.

\bibitem{finn_meta-learning_2017}
Chelsea Finn and Sergey Levine.
\newblock Meta-learning and universality: Deep representations and gradient
  descent can approximate any learning algorithm.
\newblock In {\em International Conference on Learning Representations}, 2018.

\bibitem{finn_probabilistic_2018}
Chelsea Finn, Kelvin Xu, and Sergey Levine.
\newblock Probabilistic {Model}-{Agnostic} {Meta}-{Learning}.
\newblock In {\em Advances in Neural Information Processing Systems}, 2018.

\bibitem{grant_recasting_2018}
Erin Grant, Chelsea Finn, Sergey Levine, Trevor Darrell, and Thomas Griffiths.
\newblock Recasting gradient-based meta-learning as hierarchical bayes.
\newblock In {\em International Conference on Learning Representations}, 2018.

\bibitem{gu2017deep}
Shixiang Gu, Ethan Holly, Timothy Lillicrap, and Sergey Levine.
\newblock Deep reinforcement learning for robotic manipulation with
  asynchronous off-policy updates.
\newblock In {\em International Conference on Robotics and Automation}, 2017.

\bibitem{haarnoja18b}
Tuomas Haarnoja, Aurick Zhou, Pieter Abbeel, and Sergey Levine.
\newblock Soft actor-critic: Off-policy maximum entropy deep reinforcement
  learning with a stochastic actor.
\newblock In {\em International Conference on Machine Learning}, 2018.

\bibitem{he2016deep}
Kaiming He, Xiangyu Zhang, Shaoqing Ren, and Jian Sun.
\newblock Deep residual learning for image recognition.
\newblock In {\em IEEE Conference on Computer Vision and Pattern Recognition},
  2016.

\bibitem{hu2018synthesized}
Hexiang Hu, Liyu Chen, Boqing Gong, and Fei Sha.
\newblock Synthesized policies for transfer and adaptation across tasks and
  environments.
\newblock In {\em Neural Information Processing Systems}, 2018.

\bibitem{hu2018squeeze}
Jie Hu, Li~Shen, and Gang Sun.
\newblock Squeeze-and-excitation networks.
\newblock In {\em IEEE Conference on Computer Vision and Pattern Recognition},
  2018.

\bibitem{huh2019feedback}
Minyoung Huh, Shao-Hua Sun, and Ning Zhang.
\newblock Feedback adversarial learning: Spatial feedback for improving
  generative adversarial networks.
\newblock In {\em IEEE Conference on Computer Vision and Pattern Recognition},
  2019.

\bibitem{kalashnikov18a}
Dmitry Kalashnikov, Alex Irpan, Peter Pastor, Julian Ibarz, Alexander Herzog,
  Eric Jang, Deirdre Quillen, Ethan Holly, Mrinal Kalakrishnan, Vincent
  Vanhoucke, and Sergey Levine.
\newblock Scalable deep reinforcement learning for vision-based robotic
  manipulation.
\newblock In {\em Conference on Robot Learning}, 2018.

\bibitem{karras2019style}
Tero Karras, Samuli Laine, and Timo Aila.
\newblock A style-based generator architecture for generative adversarial
  networks.
\newblock In {\em IEEE Conference on Computer Vision and Pattern Recognition},
  2019.

\bibitem{kim_bayesian_2018}
Taesup Kim, Jaesik Yoon, Ousmane Dia, Sungwoong Kim, Yoshua Bengio, and Sungjin
  Ahn.
\newblock Bayesian model-agnostic meta-learning.
\newblock In {\em Advances in Neural Information Processing Systems}, 2018.

\bibitem{kingma2014adam}
Diederik~P Kingma and Jimmy Ba.
\newblock Adam: A method for stochastic optimization.
\newblock In {\em International Conference on Learning Representations}, 2015.

\bibitem{koch2015siamese}
Gregory Koch, Richard Zemel, and Ruslan Salakhutdinov.
\newblock Siamese neural networks for one-shot image recognition.
\newblock In {\em Deep Learning Workshop at International Conference on Machine
  Learning}, 2015.

\bibitem{omniglot}
Brenden Lake, Ruslan Salakhutdinov, Jason Gross, and Joshua Tenenbaum.
\newblock One shot learning of simple visual concepts.
\newblock In {\em Conference of the Cognitive Science Society}, 2011.

\bibitem{lee_gradient-based_2018}
Yoonho Lee and Seungjin Choi.
\newblock Gradient-{Based} {Meta}-{Learning} with {Learned} {Layerwise}
  {Metric} and {Subspace}.
\newblock In {\em International Conference on Machine Learning}, 2018.

\bibitem{lee2018gradient}
Yoonho Lee and Seungjin Choi.
\newblock Gradient-based meta-learning with learned layerwise metric and
  subspace.
\newblock In {\em International Conference on Machine Learning}, 2018.

\bibitem{lee2019composing}
Youngwoon Lee, Shao-Hua Sun, Sriram Somasundaram, Edward~S. Hu, and Joseph~J.
  Lim.
\newblock Composing complex skills by learning transition policies.
\newblock In {\em International Conference on Learning Representations}, 2019.

\bibitem{li_learning_2016}
Ke~Li and Jitendra Malik.
\newblock Learning to {Optimize}.
\newblock In {\em International Conference on Learning Representations}, 2016.

\bibitem{LillicrapHPHETS15}
Timothy~P. Lillicrap, Jonathan~J. Hunt, Alexander Pritzel, Nicolas Heess, Tom
  Erez, Yuval Tassa, David Silver, and Daan Wierstra.
\newblock Continuous control with deep reinforcement learning.
\newblock In {\em International Conference on Learning Representations}, 2016.

\bibitem{maaten2008visualizing}
Laurens van~der Maaten and Geoffrey Hinton.
\newblock Visualizing data using t-sne.
\newblock In {\em Journal of Machine Learning Research}, 2008.

\bibitem{dataset-aircraft}
Subhransu Maji, Esa Rahtu, Juho Kannala, Matthew Blaschko, and Andrea Vedaldi.
\newblock Fine-grained visual classification of aircraft.
\newblock {\em arXiv preprint airxiv:1306.5151}, 2013.

\bibitem{mishra2018a}
Nikhil Mishra, Mostafa Rohaninejad, Xi~Chen, and Pieter Abbeel.
\newblock A simple neural attentive meta-learner.
\newblock In {\em International Conference on Learning Representations}, 2018.

\bibitem{mnih2014recurrent}
Volodymyr Mnih, Nicolas Heess, Alex Graves, and koray kavukcuoglu.
\newblock Recurrent models of visual attention.
\newblock In {\em Advances in Neural Information Processing Systems}. 2014.

\bibitem{munkhdalai17a}
Tsendsuren Munkhdalai and Hong Yu.
\newblock Meta networks.
\newblock In {\em International Conference on Machine Learning}, 2017.

\bibitem{nichol2018reptile}
Alex Nichol and John Schulman.
\newblock Reptile: a scalable metalearning algorithm.
\newblock {\em arXiv preprint arXiv:1803.02999}, 2018.

\bibitem{oreshkin_tadam:_2018}
Boris~N. Oreshkin, Pau Rodriguez, and Alexandre Lacoste.
\newblock {TADAM}: {Task} dependent adaptive metric for improved few-shot
  learning.
\newblock In {\em Advances in Neural Information Processing Systems}, 2018.

\bibitem{park2019semantic}
Taesung Park, Ming-Yu Liu, Ting-Chun Wang, and Jun-Yan Zhu.
\newblock Semantic image synthesis with spatially-adaptive normalization.
\newblock In {\em IEEE Conference on Computer Vision and Pattern Recognition},
  2019.

\bibitem{perez2017learning}
Ethan Perez, Harm De~Vries, Florian Strub, Vincent Dumoulin, and Aaron
  Courville.
\newblock Learning visual reasoning without strong priors.
\newblock 2017.

\bibitem{perez_film:_2017}
Ethan Perez, Florian Strub, Harm de~Vries, Vincent Dumoulin, and Aaron
  Courville.
\newblock {FiLM}: {Visual} {Reasoning} with a {General} {Conditioning} {Layer}.
\newblock In {\em Association for the Advancement of Artificial Intelligence},
  2018.

\bibitem{peters2018deep}
Matthew Peters, Mark Neumann, Mohit Iyyer, Matt Gardner, Christopher Clark,
  Kenton Lee, and Luke Zettlemoyer.
\newblock Deep contextualized word representations.
\newblock In {\em North American Chapter of the Association for Computational
  Linguistics}, 2018.

\bibitem{Rajeswaran18}
Aravind Rajeswaran*, Vikash Kumar*, Abhishek Gupta, Giulia Vezzani, John
  Schulman, Emanuel Todorov, and Sergey Levine.
\newblock {Learning Complex Dexterous Manipulation with Deep Reinforcement
  Learning and Demonstrations}.
\newblock In {\em Robotics: Science and Systems (RSS)}, 2018.

\bibitem{ravi_optimization_2017}
Sachin Ravi and Hugo Larochelle.
\newblock Optimization as a {Model} for {Few}-{Shot} {Learning}.
\newblock In {\em International Conference on Learning Representations}, 2017.

\bibitem{rothfuss2019promp}
Jonas Rothfuss, Dennis Lee, Ignasi Clavera, Tamim Asfour, Pieter Abbeel,
  Dmitriy Shingarey, Lukas Kaul, Tamim Asfour, C~Dometios Athanasios, You Zhou,
  et~al.
\newblock Promp: Proximal meta-policy search.
\newblock In {\em International Conference on Learning Representations}, 2019.

\bibitem{rusu2018metalearning}
Andrei~A. Rusu, Dushyant Rao, Jakub Sygnowski, Oriol Vinyals, Razvan Pascanu,
  Simon Osindero, and Raia Hadsell.
\newblock Meta-learning with latent embedding optimization.
\newblock In {\em International Conference on Learning Representations}, 2019.

\bibitem{santoro_meta-learning_2016}
Adam Santoro, Sergey Bartunov, Matthew Botvinick, Daan Wierstra, and Timothy
  Lillicrap.
\newblock Meta-learning with memory-augmented neural networks.
\newblock In {\em International Conference on Machine Learning}, 2016.

\bibitem{schmidhuber:1987:srl}
Jurgen Schmidhuber.
\newblock Evolutionary principles in self-referential learning. (on learning
  how to learn: The meta-meta-... hook.).
\newblock Diploma thesis, 1987.

\bibitem{schmidhuber1998reinforcement}
J{\"u}rgen Schmidhuber, Jieyu Zhao, and Nicol~N Schraudolph.
\newblock Reinforcement learning with self-modifying policies.
\newblock In {\em Learning to learn}. 1998.

\bibitem{schmidhuber1997shifting}
J{\"u}rgen Schmidhuber, Jieyu Zhao, and Marco Wiering.
\newblock Shifting inductive bias with success-story algorithm, adaptive levin
  search, and incremental self-improvement.
\newblock {\em Machine Learning}, 1997.

\bibitem{shyam2017attentive}
Pranav Shyam, Shubham Gupta, and Ambedkar Dukkipati.
\newblock Attentive recurrent comparators.
\newblock In {\em International Conference on Machine Learning}, 2017.

\bibitem{snell2017prototypical}
Jake Snell, Kevin Swersky, and Richard Zemel.
\newblock Prototypical networks for few-shot learning.
\newblock In {\em Advances in Neural Information Processing Systems}. 2017.

\bibitem{Sung_2018_CVPR}
Flood Sung, Yongxin Yang, Li~Zhang, Tao Xiang, Philip~H.S. Torr, and Timothy~M.
  Hospedales.
\newblock Learning to compare: Relation network for few-shot learning.
\newblock In {\em IEEE Conference on Computer Vision and Pattern Recognition},
  2018.

\bibitem{thrun2012learning}
Sebastian Thrun and Lorien Pratt.
\newblock {\em Learning to learn}.
\newblock Springer Science \& Business Media, 2012.

\bibitem{todorov2012mujoco}
Emanuel Todorov, Tom Erez, and Yuval Tassa.
\newblock Mujoco: A physics engine for model-based control.
\newblock In {\em International Conference On Intelligent Robots and Systems},
  2012.

\bibitem{metadataset}
Eleni Triantafillou, Tyler Zhu, Vincent Dumoulin, Pascal Lamblin, Kelvin Xu,
  Ross Goroshin, Carles Gelada, Kevin Swersky, Pierre-Antoine Manzagol, and
  Hugo Larochelle.
\newblock Meta-dataset: A dataset of datasets for learning to learn from few
  examples.
\newblock In {\em Meta-Learning Workshop at Neural Information Processing
  Systems}, 2018.

\bibitem{vaswani_attention_2017}
Ashish Vaswani, Noam Shazeer, Niki Parmar, Jakob Uszkoreit, Llion Jones,
  Aidan~N. Gomez, Lukasz Kaiser, and Illia Polosukhin.
\newblock Attention is all you need.
\newblock In {\em Advances in Neural Information Processing Systems}, 2017.

\bibitem{vinyals2016matching}
Oriol Vinyals, Charles Blundell, Tim Lillicrap, Daan Wierstra, et~al.
\newblock Matching networks for one shot learning.
\newblock In {\em Advances in Neural Information Processing Systems}, 2016.

\bibitem{vuorio2018meta}
Risto Vuorio, Dong-Yeon Cho, Daejoong Kim, and Jiwon Kim.
\newblock Meta continual learning.
\newblock {\em arXiv preprint arXiv:1806.06928}, 2018.

\bibitem{dataset-bird}
Catherine Wah, Steve Branson, Peter Welinder, Pietro Perona, and Serge
  Belongie.
\newblock The caltech-ucsd birds-200-2011 dataset.
\newblock 2011.

\bibitem{wang2018prefrontal}
Jane~X Wang, Zeb Kurth-Nelson, Dharshan Kumaran, Dhruva Tirumala, Hubert Soyer,
  Joel~Z Leibo, Demis Hassabis, and Matthew Botvinick.
\newblock Prefrontal cortex as a meta-reinforcement learning system.
\newblock {\em Nature neuroscience}, 2018.

\bibitem{wang2016learning}
Jane~X Wang, Zeb Kurth-Nelson, Dhruva Tirumala, Hubert Soyer, Joel~Z Leibo,
  Remi Munos, Charles Blundell, Dharshan Kumaran, and Matt Botvinick.
\newblock Learning to reinforcement learn.
\newblock {\em arXiv preprint arXiv:1611.05763}, 2016.

\bibitem{xie2018attentional}
Saining Xie, Sainan Liu, Zeyu Chen, and Zhuowen Tu.
\newblock Attentional shapecontextnet for point cloud recognition.
\newblock In {\em IEEE Conference on Computer Vision and Pattern Recognition},
  2018.

\bibitem{xu2015show}
Kelvin Xu, Jimmy Ba, Ryan Kiros, Kyunghyun Cho, Aaron Courville, Ruslan
  Salakhudinov, Rich Zemel, and Yoshua Bengio.
\newblock Show, attend and tell: Neural image caption generation with visual
  attention.
\newblock In {\em International Conference on Machine Learning}, 2015.

\bibitem{yang2016stacked}
Zichao Yang, Xiaodong He, Jianfeng Gao, Li~Deng, and Alex Smola.
\newblock Stacked attention networks for image question answering.
\newblock In {\em IEEE Conference on Computer Vision and Pattern Recognition},
  2016.

\bibitem{YeHZS2018Learning}
Han-Jia Ye, Hexiang Hu, De-Chuan Zhan, and Fei Sha.
\newblock Learning embedding adaptation for few-shot learning.
\newblock {\em arXiv preprint arXiv:1812.03664}, 2018.

\bibitem{zhang_self-attention_2018}
Han Zhang, Ian Goodfellow, Dimitris Metaxas, and Augustus Odena.
\newblock Self-attention generative adversarial networks.
\newblock In {\em International Conference on Machine Learning}, 2019.

\end{thebibliography}
